\documentclass[journal]{IEEEtran}
\usepackage{mathrsfs}
\usepackage{multirow}
\usepackage{booktabs}
\usepackage{hyperref}
\usepackage{color}
\usepackage{algorithm} 
\usepackage{algorithmic} 

\usepackage{flushend}

\usepackage{colortbl}
\definecolor{indian1}{rgb}{0,0,.6125}
\definecolor{indian2}{rgb}{0,0.0769,1}
\definecolor{indian3}{rgb}{0,0.6154,1}
\definecolor{indian4}{rgb}{0,0.0769,1}
\definecolor{indian5}{rgb}{0.5385,1,0.9231}
\definecolor{indian6}{rgb}{1,0.9230,0}
\definecolor{indian7}{rgb}{1,0.4615,0}
\definecolor{indian8}{rgb}{0.923,0,0}
\definecolor{indian9}{rgb}{0,0.2308,0}
\definecolor{indian10}{rgb}{0,	0.692307692307692,	0}
\definecolor{indian11}{rgb}{0.153846153846154,	0,	0.846153846153846}
\definecolor{indian12}{rgb}{0.615384615384615,	0,	0.384615384615385}
\definecolor{indian13}{rgb}{0,	0,	1}
\definecolor{indian14}{rgb}{0.538461538461538,	0,	1}
\definecolor{indian15}{rgb}{1,	0,	1}
\definecolor{indian16}{rgb}{1,	0,	0.538461538461538}

\definecolor{paviaU1}{rgb}{0,	0,	0.571428571428571}
\definecolor{paviaU2}{rgb}{0,	0.428571428571429,	1}
\definecolor{paviaU3}{rgb}{0.428571428571429,	1,	0.571428571428571}
\definecolor{paviaU4}{rgb}{1,	0.714285714285714,	0}
\definecolor{paviaU5}{rgb}{0.857142857142857,	0,	0}
\definecolor{paviaU6}{rgb}{0,	0.714285714285714,	0}
\definecolor{paviaU7}{rgb}{0.428571428571429,	0,	0.571428571428571}
\definecolor{paviaU8}{rgb}{0.142857142857143,	0,	1}
\definecolor{paviaU9}{rgb}{1,	0,	1}

\ifCLASSINFOpdf
  \usepackage[pdftex]{graphicx}
  \graphicspath{{../pdf/}{../jpeg/}}
\else
  \usepackage[dvips]{graphicx}
  \graphicspath{{../eps/}}
\fi

\usepackage{algorithm} 
\usepackage{algorithmic} 
\usepackage{amsmath,amsthm,amssymb}

\begin{document}
%
\title{Hyperspectral Image Classification in the Presence of Noisy Labels}
%
%
%

\author{Junjun~Jiang,~\IEEEmembership{Member,~IEEE,~}
        Jiayi~Ma,~\IEEEmembership{Member,~IEEE,~}
        Zheng Wang,
        Chen Chen,~\IEEEmembership{Member,~IEEE,~}
        Xianming Liu,~\IEEEmembership{Member,~IEEE}

\thanks{The research was supported by the National Natural Science Foundation of China (61501413, 61503288, 61773295, 61672193). (\emph{Corresponding author: Jiayi Ma}).} %
\IEEEcompsocitemizethanks{


\IEEEcompsocthanksitem J. Jiang and X. Liu are with the School of Computer Science and Technology, Harbin Institute of Technology, Harbin 150001, China, and are also with Peng Cheng Laboratory, Shenzhen, China. E-mail: junjun0595@163.com; csxm@hit.edu.cn.
\IEEEcompsocthanksitem J. Ma is with the Electronic Information School, Wuhan University, Wuhan 430072, China. E-mail: jyma2010@gmail.com.
\IEEEcompsocthanksitem Z. Wang is with the Digital Content and Media Sciences Research Division, National Institute of Informatics, Tokyo 101-8430, Japan. E-mail: wangz@nii.ac.jp.
\IEEEcompsocthanksitem C. Chen is with the Center for Research in Computer Vision, University of Central Florida (UCF), Orlando, FL 32816-2365 USA. E-Mail: chenchen870713@gmail.com. 
}
\thanks{Copyright (c) 2016 IEEE. Personal use of this material is permitted. However, permission to use this material for any other purposes must be obtained from the IEEE by sending an email to pubs-permissions@ieee.org.}
}

%
%

\markboth{IEEE Transactions on Geoscience and Remote Sensing}%
{Shell \MakeLowercase{\textit{\emph{et al.}}}: Bare Demo of IEEEtran.cls for Journals}
%



\maketitle

\begin{abstract}
Label information plays an important role in supervised hyperspectral image classification problem. However, current classification methods all ignore an important and inevitable problem---labels may be corrupted and collecting clean labels for training samples is difficult, and often impractical. Therefore, how to learn from the database with noisy labels is a problem of great practical importance. In this paper, we study the influence of label noise on hyperspectral image classification, and develop a random label propagation algorithm (RLPA) to cleanse the label noise. The key idea of RLPA is to exploit knowledge (\emph{e.g.}, the superpixel based spectral-spatial constraints) from the observed hyperspectral images and apply it to the process of label propagation. Specifically, RLPA first constructs a spectral-spatial probability transfer matrix (SSPTM) that simultaneously considers the spectral similarity and superpixel based spatial information. It then randomly chooses some training samples as ``clean'' samples and sets the rest as unlabeled samples, and propagates the label information from the ``clean'' samples to the rest unlabeled samples with the SSPTM. By repeating the random assignment (of ``clean'' labeled samples and unlabeled samples) and propagation, we can obtain multiple labels for each training sample. Therefore, the final propagated label can be calculated by a majority vote algorithm. Experimental studies show that RLPA can reduce the level of noisy label and demonstrates the advantages of our proposed method over four major classifiers with a significant margin---the gains in terms of the average OA, AA, Kappa are impressive, \emph{e.g.}, 9.18\%, 9.58\%, and 0.1043. \textcolor[rgb]{1.00,0.00,0.00}{The Matlab source code is available at \url{https://github.com/junjun-jiang/RLPA}}.
\end{abstract}


\section{Introduction}
\label{sec:intro}
\IEEEPARstart{D}ue to the rapid development and proliferation of hyperspectral remote sensing technology, hundreds of narrow spectral wavelengths for each image pixel can be easily acquired by space borne or airborne sensors, such as AVIRIS, HyMap, HYDICE, and Hyperion.
This detailed spectral reflectance signature makes accurately discriminating materials of interest possible~\cite{brown2006spectral, fauvel2013advances, ma2018guided}. Because of the numerous demands in ecological science, ecology management, precision agriculture, and military applications, a large number of hyperspectral image classification algorithms have appeared on the scene \cite{jia2013feature, brown2008marte, brown2010hydrothermal, he2018recent} by exploiting the spectral similarity and spectral-spatial feature \cite{jiang2017spatial, ji2014spectral, kang2014spectral, Tong2017RS}. These methods can be divided into two categories: supervised and unsupervised. The former is generally based on clustering first and then manually determining the classes. Through incorporating the label information, these supervised methods leverage powerful machine learning algorithms to train a decision rule to predict the labels of the testing pixels. In this paper, we mainly focus on the supervised hyperspectral image classification techniques.

In the past decade, the remote sensing community has introduced intensive works to establish an accurate hyperspectral image classifier. A number of supervised hyperspectral image classification methods have been proposed, such as Bayesian models \cite{landgrebe2005signal}, neural networks \cite{ratle2010semisupervised}, random forest \cite{ham2005investigation, xia2014hyperspectral}, support vector machine (SVM) \cite{melgani2004classification}, sparse representation classification \cite{chen2011hyperspectral, gao2017semi}, extreme learning machine (ELM) \cite{li2015local, samat2014rm}, and their variants \cite{waske2010sensitivity}.
Benefiting from elaborately established hyperspectral image databases, these well-trained classifiers have achieved remarkably good results in terms of classification accuracy.

However, actual hyperspectral image data inevitably contain considerable noise \cite{pelletier2017effect}: feature noise and label noise. To deal with the feature noise, which is caused by limited light in individual bands, and atmospheric and instrumental factors, many spectral feature noise robust approaches have been proposed \cite{othman2006noise, prasad2012information, yuan2012hyperspectral, li2015hyperspectral}.
Label noise has received less attention than feature noise, however, it is pervasive due to the following reasons:
(i) When the information provided to an expert is very limited or the land cover is highly complex, \emph{e.g.}, low inter-class and high intra-class variabilities, it is very easy to cause mislabeling.
(ii) The low-cost, easy-to-get automatic labeling systems or inexperienced personnel assessments are less reliable \cite{snow2008cheap}.
(iii) If multiple experts label the same image at the same time, the labeling results may be inconsistent between different experts \cite{raykar2010learning}.
(iv) Information loss (due to data encoding and decoding and data dissemination) will also cause label noise.

\begin{figure*}[t]
\centering
\includegraphics[width=15.70cm]{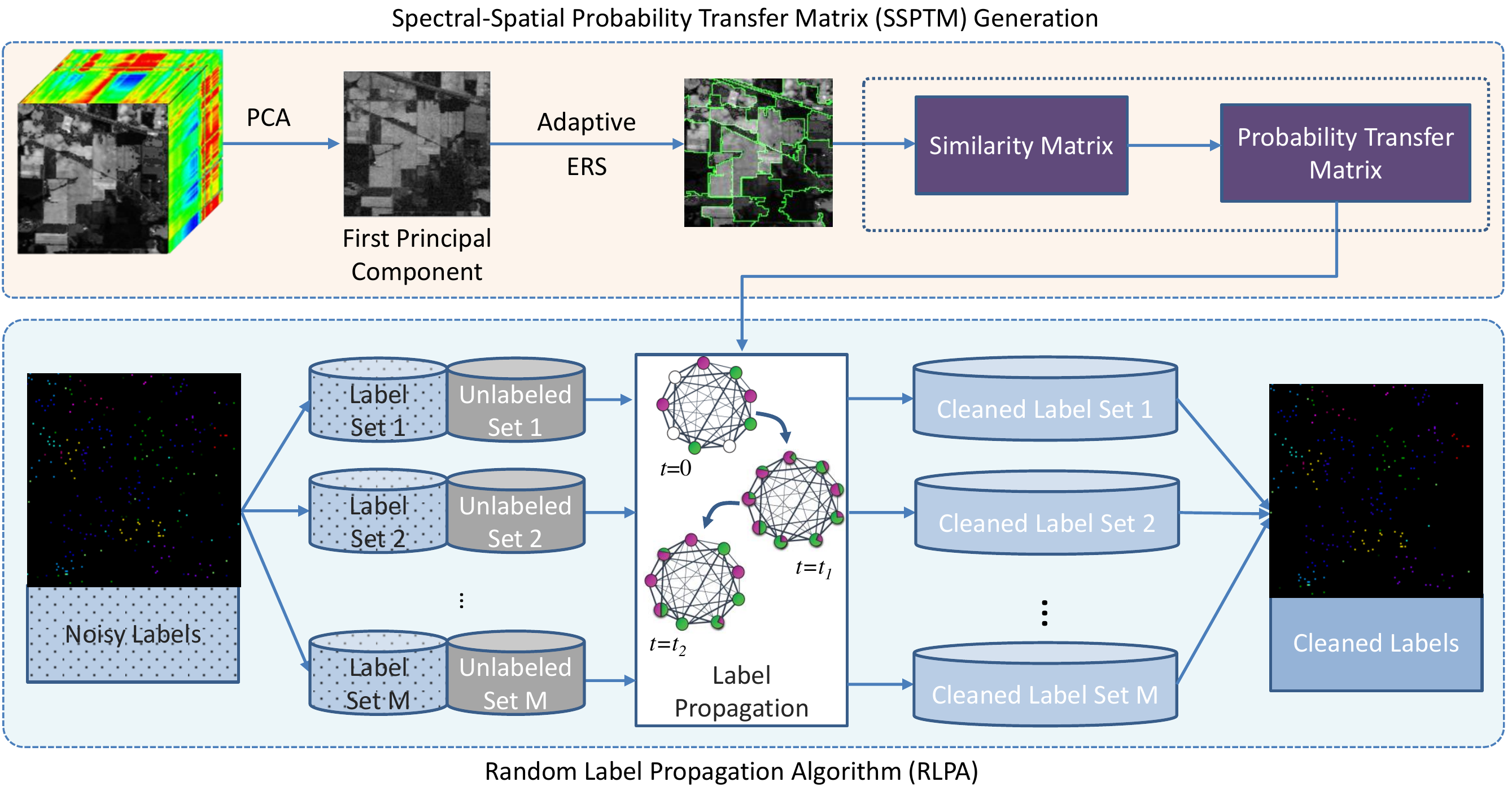}
\vspace{-0.20cm}
\caption{Schematic diagram of the proposed random label propagation algorithm based label noise cleansing process. The up dashed block demonstrates the procedure of SSPTM generation, while the bottom dashed block demonstrates the main steps of the random label propagation algorithm.}
\label{fig:diagram}
\end{figure*}

Recently, the classification problem in the presence of label noise is becoming increasingly important and many label noise robust classification algorithms have been proposed \cite{angluin1988learning, lawrence2001estimating, natarajan2013learning, liu2016classification}. These methods be divided into two major categories: label noise-tolerant classification and label noise cleansing.
The former adopts the strategies of bagging and boosting, or decision tree based ensemble techniques, while the latter aims to filter the label noise by exploiting the prior knowledge of the training samples. For more details about the general classification problem with label noise, interested reader is referred to \cite{frenay2014classification} and the references therein. Generally speaking, the label noise-tolerant classification model is often designed for a specific classifier, so that the algorithm lacks universality. In contrast, as a pre-processing method, the label noise cleansing method is more general and can be used for any classifier, including the above-mentioned noisy label robust classification model. Therefore, this study will focus on the more universal noisy label cleansing approach.

Although considerable literature deals with the general image classification, there is very little research work on the classification of hyperspectral images under noisy labels \cite{pelletier2017effect, condessa2016supervised}. However, in the actual classification of hyperspectral images, this is a more urgent and unavoidable problem. As reported by Pelletier \emph{et al}.'s study \cite{pelletier2017effect}, the noisy labels will mislead the training procedure of the hyperspectral image classification algorithm and severely decrease the classification accuracy of land cover. Nevertheless, there is still relatively little work specifically developed for hyperspectral image classification when encountered with label noise. Therefore, hyperspectral image classification in the presence of noisy labels is a problem that requires a solution.

In this paper, we propose to exploit the spectral-spatial constraints based knowledge to guide the cleansing of noisy labels under the label propagation framework. In particular, we develop a random label propagation algorithm (RLPA). As shown in Fig. \ref{fig:diagram}, it includes two steps: (i) spectral-spatial probability transfer matrix (SSPTM) generation and (ii) random label propagation.
At the first step, considering that spatial information is very important for the similarity measurement of different pixels \cite{ji2014spectral, Pu2014Novel, kang2014spectral, zheng2017dimensionality}, we propose a novel affinity graph construction method which simultaneously considers the spectral similarity and the superpixel segmentation based spatial constraint. The SSPTM can be generated through the constructed affinity graph.
In the second step, we randomly divide the training database to a labeled subset (with ``clean'' labels) and an unlabeled subset (without labels), and then perform the label propagation procedure on the affinity graph to propagate the label information from labeled subset to the unlabeled subset. Since the process of random assignment (of ¡°clean¡± labeled
samples and unlabeled samples) and propagation can be executed multiple times, the unlabeled subset will receive the multiple propagated labels. Through fusing the multiple labels of many label propagation steps with a majority vote algorithm (MVA), it can be expected to cleanse the label information. The philosophy behind this is that the samples with real labels dominate all training classes, and we can gradually propagate the clean label information to the entire dataset by random splitting and propagation.
The proposed method is tested on three real hyperspectral image databases, namely the Indian Pines, University of Pavia, and Salinas Scene, and compared to some existing approaches using overall accuracy (OA), average accuracy (AA), and
the kappa metrics. It is shown that the proposed method outperforms these methods in terms of objective metrics and visual classification map.

The main contributions of this article can be summarized as follows:
\begin{itemize}
  \item We provide an effective solution for hyperspectral image classification in the presence of noisy labels. It is very general and can be seamlessly applied to the current classifiers.
  \item By exploiting the hyperspectral image prior, \emph{i.e.}, the superpixel based spectral-spatial constraints, we propose a novel probability transfer matrix generation method, which can ensure label information of the same class propagate to each other, and prevent the label propagation of samples from different classes.
  \item The proposed RLPA method is very effective in cleansing the label noise. Through the preprocess of RLPA, it can greatly improve the performance of the original classifiers, especially when the label noise level is very large.

\end{itemize}

This paper is organized as follows: In Section \ref{sec:setup}, we present the problem setup. Section \ref{sec:Influence} shows the influence of label noise on the hyperspectral image classification performance. In Section \ref{sec:Proposed}, the details of the proposed RLPA method are given. Simulations and experiments are presented in Section \ref{sec:Experiments}, and Section \ref{sec:Conclusion} concludes this paper.

\begin{figure*}[t]
\centering
\includegraphics[width=18.0cm]{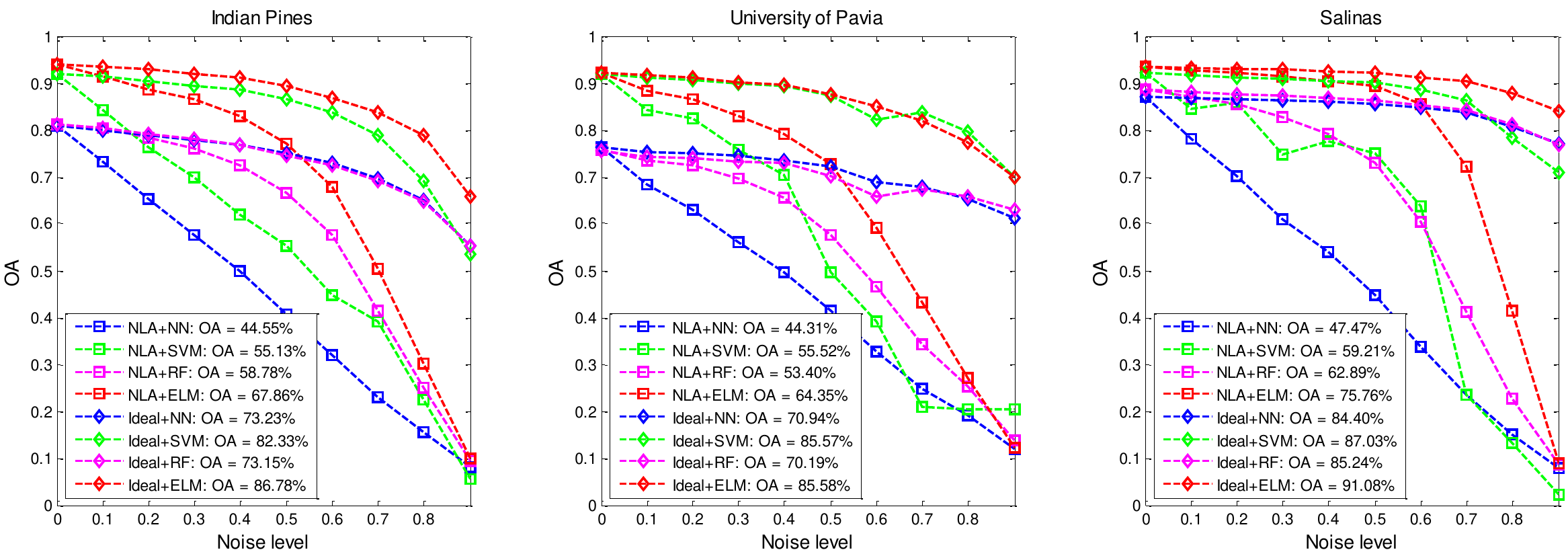}
\vspace{-0.20cm}
\caption{Influence of the label noise on the performance (in term of OA of different classifiers) on the Indian Pines, University of Pavia, and Salinas Scene databases.}
\label{fig:NoisyAnalysis}
\end{figure*}

\section{Problem Formulation}
\label{sec:setup}
In this section, we formalize the foundational definitions and setup of the noisy label hyperspectral image classification problem. A hyperspectral image cube consists of hundreds of nearly contiguous spectral bands, with high spectral resolution (5-10 nm), from the visible to infrared spectrum for each image pixel. Given some labeled pixels in a hyperspectral image, the task of hyperspectral image classification is to predict the labels of unseen pixels. Specifically, let ${\mathcal{X}=}\left\{ {{\textbf{x}}_1}, {{\textbf{x}}_2} \cdots {\rm{,}} {{\textbf{x}}_N}\right\} \in {\mathbb{R}^{D}}$ denote a database of pixels in a $D$ dimensional input spectral space, and $\mathcal{Y}=\left\{1, 2, \cdots {\rm{,}} C\right\}$ denote a label set. The class labels of $\left\{ {{{\textbf{x}}_1}, {{\textbf{x}}_2}, \cdots {\rm{,}} {{\textbf{x}}_{N}}} \right\}$ are denoted as $\left\{ {{{{y}}_1}, {{{y}}_2}, \cdots {\rm{,}} {{{y}}_{N}}} \right\}$. Mathematically, we use a matrix $\textbf{Y} \in {\mathbb{R}^{N\times C}}$ to represent the label, where $\textbf{Y}_{ij} =1$ if $\textbf{x}_i$ is labeled as $j$. In order to model the label noise process, we additionally introduce another variable $\tilde{\textbf{Y}} \in {\mathbb{R}^{N\times C}}$ that is used to denote the noise observed label. Let $\rho$ denotes the label noise level (also called error rate or noise rate~\cite{kalai2005boosting}) specifying the probability of one label being flipped to another, and thus $\rho_{jk}$ can be mathematically formalized as:
\begin{equation}\label{eq:probability}
\rho_{jk} = P(\tilde{\textbf{Y}}_{ik} =1|\textbf{Y}_{ij} =1), \forall  j \ne k, {\kern 3pt} {\rm{and}} {\kern 5pt} j,k \in \{ 1,2, \cdots ,C\}.
\end{equation}
For example, when $\rho = 0.3$, it means that for a pixel $\textbf{x}_i$, whose label is $j$, there is a 30\% probability to be labeled as the other class $k$ ($k \ne j$).
To help make sense of this, we give the pseudo-codes of the noisy label generation process in Algorithm \ref{algorithmic0}.
$\texttt{size}(\textbf{X})$ is a function that returns the sizes of each dimension of array $\textbf{X}$,
$\texttt{rand}(N)$ is a function that returns a random scalar drawn from the standard uniform distribution on the open interval $(0,1)$,
$\texttt{find}(\textbf{X})$ is a function that locates all nonzero elements of an array $\textbf{X}$,
and $\texttt{randperm} (N)$ is a function that returns a row vector containing a random permutation of the integers from 1 to $N$ inclusive.

In this paper, our main task it to predict the label of an unseen pixel $\textbf{x}_t$, with the training data $\textbf{X} = [{{\textbf{x}}_1}, {{\textbf{x}}_2}, \cdots {\rm{,}} {{\textbf{x}}_N}]$ and the noisy label matrix $\tilde{\textbf{Y}}$.


\begin{algorithm}[h]
\caption{\bfseries{Noisy label generation.}}
\begin{algorithmic}[1]
\STATE {\bfseries{Input}}: The clean label matrix $\textbf{Y}$ and the level of label noise $\rho$.
\STATE {\bfseries{Output}}: The noisy label matrix $\tilde{\textbf{Y}}$.
\STATE $[N, C]=\texttt{size}({\textbf{Y}})$;
\STATE $\tilde{\textbf{Y}} = \textbf{Y}$;
\STATE $\textbf{k} = \texttt{rand}(N,1)$;
\FOR{$i$ = 1 to $N$}
\IF{$\textbf{k}(i) \le \rho$}
\STATE $p = \texttt{find}({\textbf{Y}}_{i,:}=1$);
\STATE $\textbf{r} = \texttt{randperm}(C)$;
\STATE $\textbf{r}(p) = [{\kern 3pt}]$; {\kern 29pt}$\backslash \backslash $ $[{\kern 3pt}]$ is the null set.
\STATE $\tilde{\textbf{Y}}_{\textbf{r}(1),:}=1$;
\ENDIF
\ENDFOR
\end{algorithmic}
 \label{algorithmic0}
\end{algorithm}


\begin{figure*}[t]
\centering
\includegraphics[width=18.0cm]{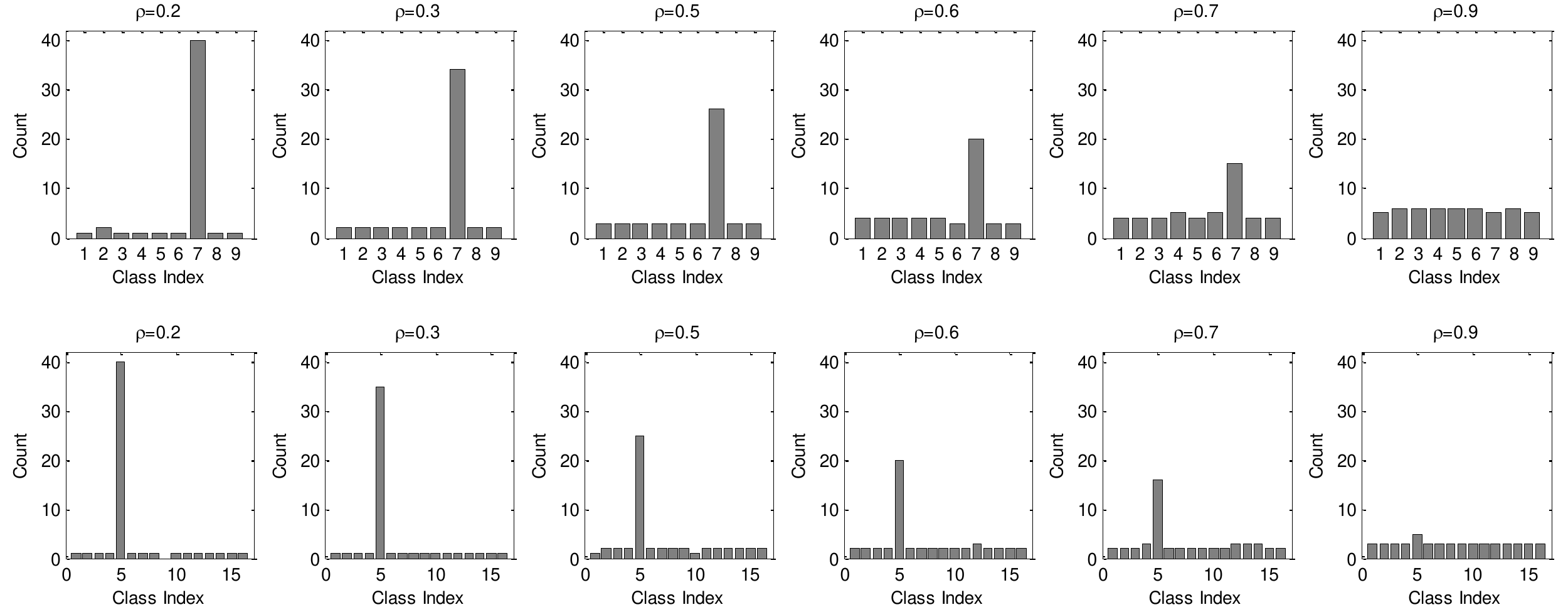}
\vspace{-0.15cm}
\caption{The distribution of a correct class at different levels of label noise $\rho$. First row: the distribution of samples with true label 7 under different label noise $\rho$ for the University of Pavia database which has nine classes. Second row: the distribution of samples with true label 5 under different levels of label noise $\rho$ for the Salinas Scene database which has 16 classes. }
\label{fig:proportion}
\end{figure*}

\section{Influence of Label Noise on Hyperspectral Image Classification}
\label{sec:Influence}
In this section, we examine the influence of label noise on the hyperspectral image classification problem. As shown in Fig. \ref{fig:NoisyAnalysis}, we demonstrate the impact of label noise on four different classifiers: neighbor nearest (NN), support vector machines (SVM), random forest (RF), and extreme learning machine (ELM). The noise level changes from 0 to 0.9 at an interval of  0.1.
In Fig. \ref{fig:NoisyAnalysis}, we report the average OA over ten runs (more details about the experimental settings can be found in Section \ref{sec:Experiments}) as a function of the noise level. Noisy label based algorithm (NLA) represents the classification with the noisy labels without cleansing. From these results, we can draw the following four conclusions:

1) With the increase of the label noise level, the performance of all classification methods is gradually declining. Meanwhile, we also notice that the impact of label noise is not identical for all classifiers. Among these four classifiers, RF and ELM are relatively robust to label noise. When the label noise level is not large, these two classifiers can obtain better performance. In contrast, NN and SVM are much more sensitive to the label noise level. The poor results of NN and SVM can be attributed to their reliance on nearest samples and support vectors.

2) The University of Pavia and Salinas Scene databases have the same number of training samples (\emph{e.g.}, 50)\footnote{In this analysis, we only paid attention to these two databases in order to avoid the impact of different numbers of training sample. For the Indian Pines database, we select 10\% samples for each class and the number of training samples is not the same as that in the two other databases.}, but the decline rate of OA on the University of Pavia is significantly faster than that of Salinas Scene database. This is mainly because that the number of classes in the Salinas database is larger than that of the University of Pavia database ($C=16$ \emph{vs}. $C=9$). With the same label noise level and same number of training samples, the more the classes are, the greater the probability of choosing the correct samples is\footnote{In this situation, for each class (after adding label noise), although the ratio of samples with corrected labels to samples with incorrectly labeled sample is $\frac{{1-\rho}}{\rho}$, this will reduce to $\frac{{1-\rho}}{\rho/(C-1)}$ when we consider the ratio of samples with corrected labels to samples labeled another class. For example, when $\rho=0.5$, $C=16$, the ratio of samples with corrected labels to samples labeled another class is 15:1.}. This point is illustrated by Fig. \ref{fig:proportion}. When the noise is not very large, \emph{e.g.}, $\rho \le 0.7$, the samples with true labels can often dominate. In this case, a good classifier can also get satisfactory performance.

3) We also show the ideal case that we know the noisy label samples and remove these training samples to obtain a noiseless training subset. From the comparisons (please refer to the same colors in each subfigure), we observe that there is considerable room of improvement for the strategy of label noise cleansing-based algorithms. This also demonstrates the importance of preprocessing based on label noise cleansing.

\section{Proposed Method}
\label{sec:Proposed}

\subsection{Overview of the Framework}
\begin{figure}[h]
\centering
\includegraphics[width=8.8cm]{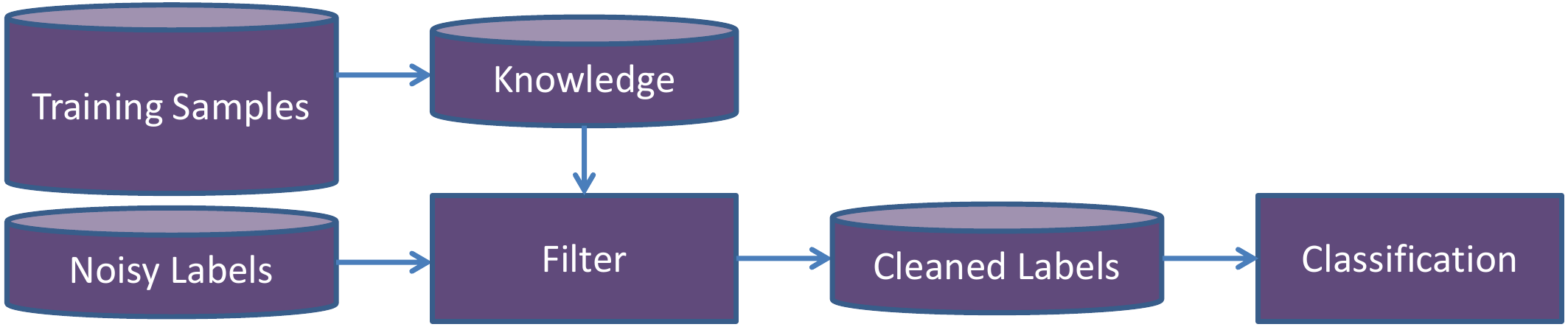}
\vspace{-0.50cm}
\caption{The typical procedure of labels cleansing  based mthod for hyperspectral image classification in the presence of label noise.}
\label{fig:cleansing}
\end{figure}

To handle the label noise, there are two main kinds of methods. The first class is to design a specific classifier that is robust to the presence of label noise, while the other obvious and tempting method is to improve the label quality of training samples. Since the latter is intuitive and can be applied to any of the subsequent classifiers, in this paper we mainly focus on how to improve and cleanse the labels. The main steps are illustrated by Fig. \ref{fig:cleansing}. Firstly, the prior knowledge (\emph{e.g.}, neighborhood relationship or topology) is extracted from the training set and used to regularize the filter of label noise. Based on the cleaned labels, we can expect an intermediate classification result. 

The core idea of the proposed label cleansing approach is to randomly remove the labels of some selected samples, and then apply the label propagation algorithm to predict the labels of these selected (unlabeled) samples according to a predefined SSPTM. The philosophy behind this method is that the samples with correct labels account for the majority, therefore, we can gradually propagate the clean label information to the entire samples by random splitting and propagation. This is reasonable because when the samples with wrong labels account for the majority, we cannot obtain the clean label for the samples anyway. \textcolor[rgb]{0.00,0.00,0.00}{As we know, traditional label propagation methods are sensitive to noise. This is mainly because when the label contains noise and there is no extra prior information, it is very hard for these traditional methods to construct a reasonable probability transfer matrix. The label noise can not only be removed, but is likely to be spread. Though our method is also label propagation based, we can take full advantage of the priori knowledge of hyperspectral images, \emph{i.e.}, the superpixel based spectral-spatial constraint, to construct the SSPTM, which is the key to this label propagation based algorithm. Based on the constructed SSPTM, we can ensure that samples with same classes can be propagated to each other with a high probability, and samples with different classes cannot be propagated.}

Fig. \ref{fig:diagram} illustrates the schematic diagram of the proposed method. In the following, we will first introduce how to generate the probability transfer matrix with both the spectral and spatial constraints. Then we present the random label propagation approach.

\subsection{Construction of Spectral-Spatial Affinity Graph}
The definition of the edge weights between neighbors is the key problem in constructing an affinity graph. To measure the similarity between pixels in a hyperspectral image, the simplest way is to calculate the spectral difference through Euclidean distance, spectral angle mapper (SAM), spectral correlation mapper (SCM), or spectral information measure (SIM). However, these measurements all ignore the rich spatial information contained in a hyperspectral image, and the spectral similarity is often inaccurate due to low inter-class and high intra-class variabilities.

Our goal is to propagate label information only among samples with the same category. However, the spectral similarity based affinity graph cannot prevent label propagation of similar samples with different classes. In this paper, we propose a spectral-spatial similarity measurement approach. The basic assumption of our method is that the hyperspectral image has many homogeneous regions and pixels from one homogeneous region are more likely to be the same class. Therefore, when defining the edge weights of the affinity graph, the spectral similarity as well as the spatial constraint is taken into account at the same time.

\subsubsection{Generation of Homogeneous Regions}
As in many superpixel segmentation based hyperspectral image classification and restoration methods \cite{li2015efficient, jiang2018superpca, zhang2017multiscale, Fan2017Hyperspectral}, we adopt entropy rate superpixel segmentation (ESR) \cite{liu2011entropy} due to its promising performance in both efficiency and efficacy. Other state-of-the-art methods such as simple linear iterative clustering (SLIC) \cite{Achanta2012SLIC} can also be used to replace the ERS. Specially, we first obtain the first principal component (through principal component analysis (PCA) \cite{wold1987principal}) of hyperspectral images, $I_f$, capturing the major information of hyperspectral images. This further reduces the computational cost for superpixel segmentation. It should be noted that other state-of-the-art methods such as \cite{wang2016salient} can also be equally used to replace the PCA. Then, we perform ESR on $I_f$ to obtain the superpixel segmentation,
\begin{equation}\label{eq:segmentation}
I_f = \bigcup\limits_k^T {{\mathscr{X}_k}} ,{\kern 1pt} {\kern 1pt} {\kern 1pt} {\kern 1pt} {\kern 1pt} s.t.{\kern 1pt} {\kern 1pt} {\kern 1pt} {\mathscr{X}_k} \cap {\mathscr{X}_g} = \emptyset ,{\kern 1pt} {\kern 1pt} {\kern 1pt} (k \ne g),
\end{equation}
where $T$ denotes the number of superpixels, and $\mathscr{X}_k$ is the $k$-th superpixel. The setting of $T$ is an open and challenging problem, and is usually set experimentally. Following \cite{fang2018extinction}, we also introduce an adaptive parameter setting scheme to determine the value of $T$ by exploiting the texture information. Specifically, the Laplacian of Gaussian (LoG) operator \cite{canny1987computational} is applied to detect the image structure of the first principal component of hyperspectral images. Then we can measure the texture complexity of hyperspectral images based on the detected edge image. The more complex the texture of hyperspectral images, the larger the number of superpixels, and vice versa. Therefore, we define the number of superpixel as follows:
\begin{equation}\label{eq:ratio}
T = T_{base}\frac{{{N_f}}}{{N_I}},
\end{equation}
where $N_f$ denotes the number of nonzero elements in the detected edge image, $N_I$ is the size of $I_f$, \emph{i.e.}, the total number of pixels in $I_f$, and $T_{base}$ is a fixed number for all hyperspectral images. In this way, the number of superpixels $T$ is set adaptively, based on the spatial characteristics of different hyperspectral images. In all our experiments, we set $T_{base} = 2000$.

\begin{figure}[t]
\centering
\includegraphics[width=8.8cm]{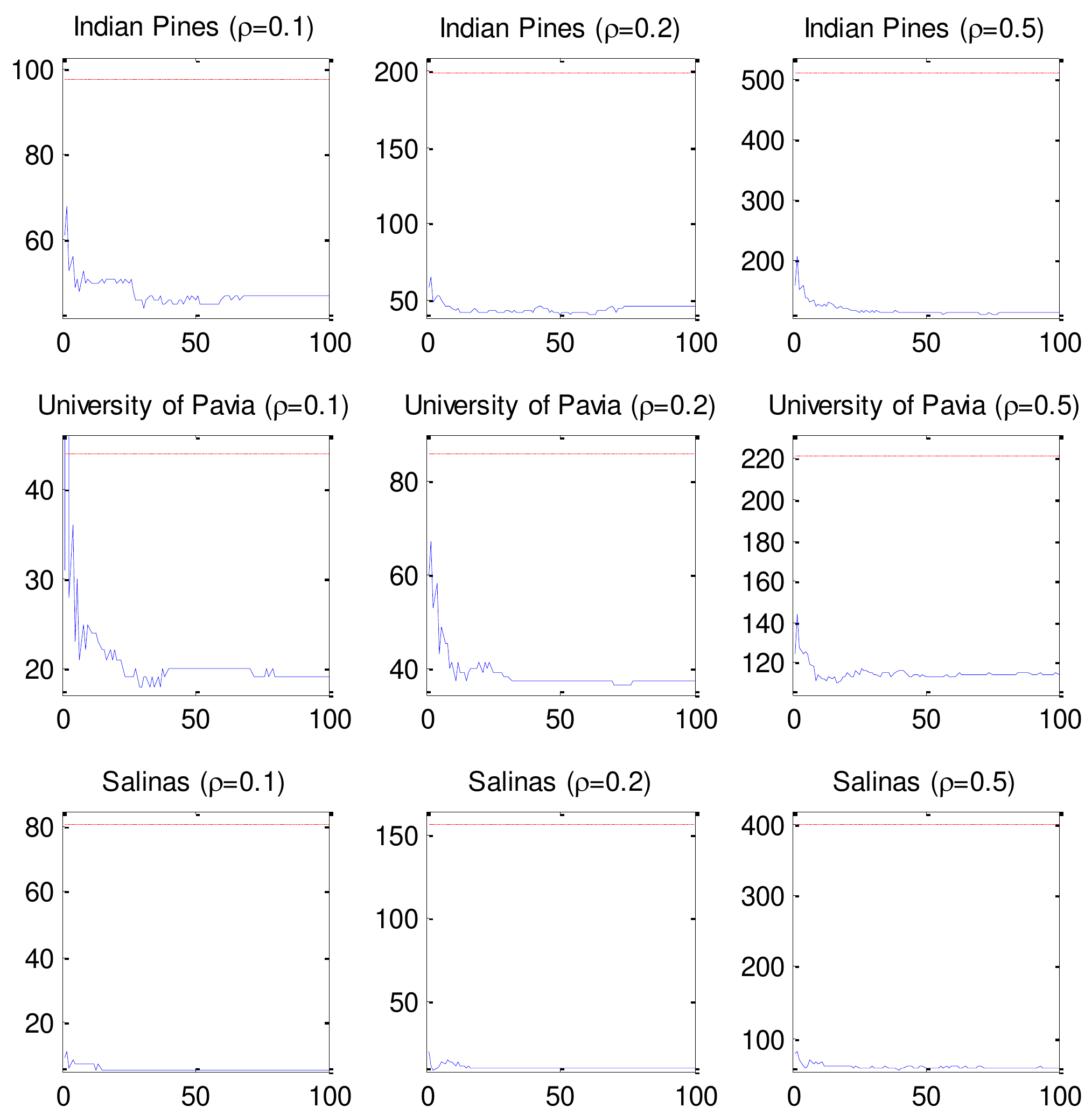}
\vspace{-0.50cm}
\caption{Plots of the number of noisy label samples (N.N.L.S.) according to the iterations of the RLPA (blue line) under three different noise levels ($\rho = 0.1, 0.2, 0.5$). We also show the initial number (red dashed line) of noisy label samples for comparison.}
\label{fig:NNLS}
\end{figure}

\subsubsection{Construction of Spectral-Spatial Regularized Probabilistic Transition Matrix}
Based on the segmentation result, we can construct the affinity graph by putting an edge between pixels within a homogeneous region and letting the edge weights between pixels from different homogeneous region be zero:
\begin{equation}\label{eq:GraphWeights}
{\textbf{W}_{ij}} = \left\{ \begin{array}{l}
\exp  \left(- {\frac{{sim{{({\textbf{x}}_i,{\textbf{x}}_j)}^2}}}{2\sigma^2 }} \right),   {\kern 13pt} {\textbf{x}}_i,{\textbf{x}}_j \in {\mathscr{X}_k}, \\
 0,   {\kern 88pt} {\textbf{x}}_i \in {\mathscr{X}_k} {\kern 5pt} {\rm{and}} {\kern 5pt} {\textbf{x}}_j \in {\mathscr{X}_g}.  \\
 \end{array} \right.
\end{equation}
Here, $sim({\textbf{x}}_i,{\textbf{x}}_j)$ denotes the spectral similarity of ${\textbf{x}}_i$ and ${\textbf{x}}_j$. In this paper, we use the Euclidean distance to measure their similarity,
\begin{equation}\label{eq:similarity}
sim({\textbf{x}}_i,{\textbf{x}}_j) = ||{\textbf{x}}_i - {\textbf{x}}_j||_2,
\end{equation}
where $\left\|  \cdot  \right\|_2$ is the $l_2$ norm of a vector. In Eq. (\ref{eq:GraphWeights}), the variance $\sigma$ is calculated region adaptively through the mean variance of all pixels in each homogeneous region:
\begin{equation}\label{eq:sigma}
\sigma = {\left( {\frac{1}{{|{\mathscr{X}_k}|}}\sum\limits_{{\textbf{x}}_i,{\textbf{x}}_j \in {\mathscr{X}_k}} {{{\left\| {{\textbf{x}}_i - {\textbf{x}}_j} \right\|}_2^2}} } \right)^{0.5}},
\end{equation}
where $\left|  \cdot  \right|$ is the cardinality operator.

Upon acquiring the spectral-spatial regularized affinity graph, the label information can be propagated between nodes through the connected edges. The larger the weight between two nodes, the easier it becomes to travel. Therefore, we can define a probability transition matrix $\textbf{T}$:
\begin{equation}\label{eq:ptm}
{\textbf{T}_{ij}} = P(j \to i) = \frac{{{\textbf{W}_{ij}}}}{{\sum\nolimits_{k = 1}^N {{\textbf{W}_{kj}}} }},
\end{equation}
where ${\textbf{T}_{ij}}$ can be seen as the probability to jump from node $j$ to node $i$.

\begin{figure*}[t]
\centering
\includegraphics[width=18.0cm]{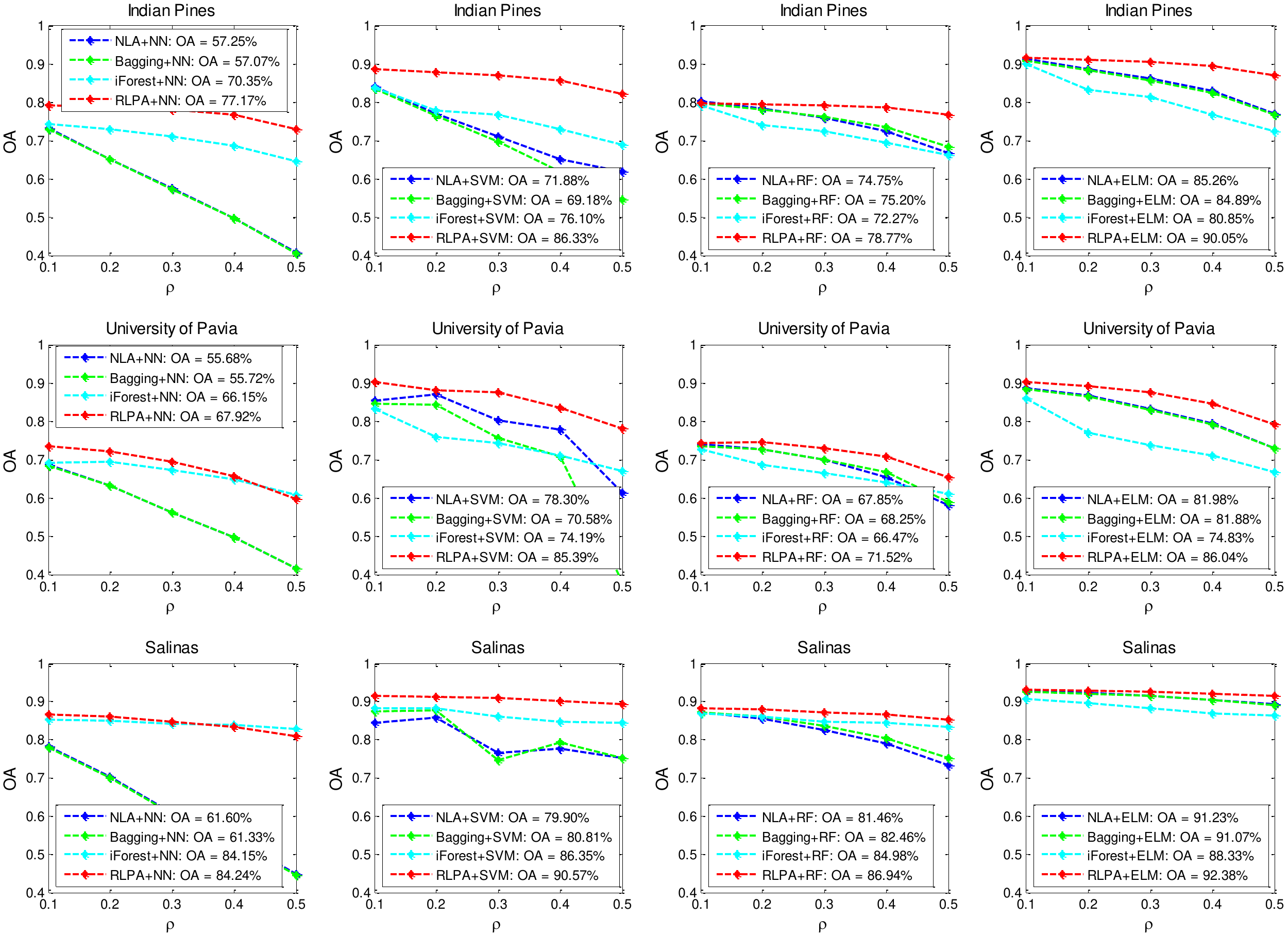}
\vspace{-0.20cm}
\caption{The quantitative classification results in term of OA of four different methods (NLA, Bagging, iForest, and RLPA) with four different classifiers (NN, SVM, RF, and ELM) on the Indian Pines (the first row), University of Pavia (the second row), and Salinas Scene (the third row). The average OAs of four different methods on three databases with four different classifiers are: NLA (OA = 73.93\%), Bagging (OA = 73.20\%), iForest (OA = 77.09\%), and RLPA (OA = 83.11\%).}
\label{fig:Improvement}
\end{figure*}

\begin{table*}[t]
\scriptsize
      \caption{Number of samples in the Indian Pines, University of Pavia, and Salinas Scene images. The background color is used to distinguish different classes.}
      \centering
        \begin{tabular}[ht]{ |c| c || c | c  || c| c| }
\hline
\multicolumn{2}{|c||}{Indian Pines}  &\multicolumn{2}{c||}{University of Pavia}  &\multicolumn{2}{c|}{Salinas Scene} \\
\hline
Class Names	&	Numbers	&	Class Names	&	Numbers	&	Class Names	&	Numbers	\\
\hline
\hline
 \multicolumn{1}{>{\columncolor{indian1}}c}{Alfalfa}	&	46	&	 \multicolumn{1}{>{\columncolor{paviaU1}}c}{Asphalt}	&	6631	&	 \multicolumn{1}{>{\columncolor{indian1}}c}{Brocoli\_green\_weeds\_1}	&	2009	\\
 \multicolumn{1}{>{\columncolor{indian2}}c}{Corn-notill}	&	1428	&	\multicolumn{1}{>{\columncolor{paviaU2}}c}{Bare soil}	&	18649	&	 \multicolumn{1}{>{\columncolor{indian2}}c}{Brocoli\_green\_weeds\_2} 	&	3726	\\
 \multicolumn{1}{>{\columncolor{indian3}}c}{Corn-mintill}	&	830	&	\multicolumn{1}{>{\columncolor{paviaU3}}c}{Bitumen}	&	2099	&	 \multicolumn{1}{>{\columncolor{indian3}}c}{Fallow} 	&	1976	\\
 \multicolumn{1}{>{\columncolor{indian4}}c}{Corn}	&	237	&	\multicolumn{1}{>{\columncolor{paviaU4}}c}{Bricks}	&	3064	&	 \multicolumn{1}{>{\columncolor{indian4}}c}{Fallow\_rough\_plow} 	&	1394	\\
 \multicolumn{1}{>{\columncolor{indian5}}c}{Grass-pasture}	&	483	&	\multicolumn{1}{>{\columncolor{paviaU5}}c}{Gravel}	&	1345	&	 \multicolumn{1}{>{\columncolor{indian5}}c}{Fallow\_smooth} 	&	2678	\\
 \multicolumn{1}{>{\columncolor{indian6}}c}{Grass-trees}	&	730	&	\multicolumn{1}{>{\columncolor{paviaU6}}c}{Meadows}	&	5029	&	 \multicolumn{1}{>{\columncolor{indian6}}c}{Stubble} 	&	3959	\\
 \multicolumn{1}{>{\columncolor{indian7}}c}{Grass-pasture-mowed}	&	28	&	\multicolumn{1}{>{\columncolor{paviaU7}}c}{Metal sheets}	&	1330	&	 \multicolumn{1}{>{\columncolor{indian7}}c}{Celery} 	&	3579	\\
 \multicolumn{1}{>{\columncolor{indian8}}c}{Hay-windrowed}	&	478	&	\multicolumn{1}{>{\columncolor{paviaU8}}c}{Shadows}	&	3682	&	 \multicolumn{1}{>{\columncolor{indian8}}c}{Grapes\_untrained} 	&	11271	\\
 \multicolumn{1}{>{\columncolor{indian9}}c}{Oats}	&	20	&		\multicolumn{1}{>{\columncolor{paviaU9}}c}{Trees} &	947	&	 \multicolumn{1}{>{\columncolor{indian9}}c}{Soil\_vinyard\_develop} 	&	6203	\\
 \multicolumn{1}{>{\columncolor{indian10}}c}{Soybean-notill}	&	972	&		&		&	 \multicolumn{1}{>{\columncolor{indian10}}c}{Corn\_senesced\_green\_weeds} 	&	3278	\\
 \multicolumn{1}{>{\columncolor{indian11}}c}{Soybean-mintill} 	&	2455	&		&		&	 \multicolumn{1}{>{\columncolor{indian11}}c}{Lettuce\_romaine\_4wk} 	&	1068	\\
 \multicolumn{1}{>{\columncolor{indian12}}c}{Soybean-clean}	&	593	&		&		&	 \multicolumn{1}{>{\columncolor{indian12}}c}{Lettuce\_romaine\_5wk} 	&	1927	\\
 \multicolumn{1}{>{\columncolor{indian13}}c}{Wheat}	&	205	&		&		&	 \multicolumn{1}{>{\columncolor{indian13}}c}{Lettuce\_romaine\_6wk} 	&	916	\\
 \multicolumn{1}{>{\columncolor{indian14}}c}{Woods}	&	1265	&		&		&	 \multicolumn{1}{>{\columncolor{indian14}}c}{Lettuce\_romaine\_7wk} 	&	1070	\\
 \multicolumn{1}{>{\columncolor{indian15}}c}{Buildings-Grass-Trees-Drives}	&	386	&		&		&	 \multicolumn{1}{>{\columncolor{indian15}}c}{Vinyard\_untrained} 	&	7268	\\
 \multicolumn{1}{>{\columncolor{indian16}}c}{Stone-Steel-Towers} 	&	93	&		&		&	 \multicolumn{1}{>{\columncolor{indian16}}c}{Vinyard\_vertical\_trellis}	&	1807	\\
\hline
Total Number&10249&Total Number&42776&Total Number&54129\\
\hline
\end{tabular}
\label{table:Three_sample}
\end{table*}

\subsection{Random Label Propagation through Spectral-Spatial Neighborhoods}
It is a very challenging problem to cleanse the label noise from the original label space. However, as a hyperspectral image, we can exploit the availableinformation about the spectral-spatial knowledge to guide the labeling of adjacent pixels. Specifically, to cleanse the noise of labels, we propose a RLPA based method. We randomly select some noisy training samples as ``clean' labeled samples and set the remaining samples as unlabeled samples. The label propagation algorithm is then used to propagate the information from the ``clean'' labeled samples to the unlabeled samples.

Concretely, we divide the training database $\mathcal{X}$ to a labeled subset $\mathcal{X}_L = \left\{ {{{\textbf{x}}_1}, {{\textbf{x}}_2}, \cdots {\rm{,}} {{\textbf{x}}_{l}}} \right\}$, whose label matrix is denoted as $\tilde{\textbf{Y}}_L =\tilde{\textbf{Y}}(:,1:l) \in {\mathbb{R}^{l\times C}}$, and an unlabeled subset $\mathcal{X}_U = \left\{ {{{\textbf{x}}_{l+1}}, {{\textbf{x}}_{l+2}}, \cdots {\rm{,}} {{\textbf{x}}_{N}}} \right\}$, whose labels are discarded.  $l$ is the number of training samples that are selected for building up the ``clean'' labeled subset, $l = \texttt{round}(N*\eta)$. Here, $\eta$ denotes the ``clean'' sample proportion in the total training samples, and $\texttt{round} (a)$  is a function that rounds the elements of $a$ to the nearest integers. It should be noted that we set the first $l$ pixels as the labeled subset, and the rest as the unlabeled subset for the convenience of expression. In our experiments, these two subsets are randomly selected from the training database $\mathcal{X}$. Now, our task is to predict the labels $\tilde{\textbf{Y}}_U$ of unlabeled pixels $\mathcal{X}_U$, based on the graph constructed by the superpixel based spectral-spatial affinity graph.

\begin{algorithm}[t]
\caption{\bfseries{Random label propagation algorithm (RLPA) based label noise cleansing.}}
\begin{algorithmic}[1]
\STATE {\bfseries{Input}}: Training samples $\left\{ {{{\textbf{x}}_1}, {{\textbf{x}}_2}, \cdots {\rm{,}} {{\textbf{x}}_{N}}} \right\}$, and the corresponding labels $\left\{ {{{{y}}_1}, {{{y}}_2}, \cdots {\rm{,}} {{{y}}_{N}}} \right\}$, parameters $\eta$ and $\alpha$.
\STATE {\bfseries{Output}}: The cleaned labels $\left\{ {{{{y}}_1^{*}}, {{{y}}_2^{*}}, \cdots {\rm{,}} {{{y}}_{N}^{*}}} \right\}$.
\FOR{$s$ = 1 to $S$}
\STATE $\texttt{rand}(`seed`,s)$;
\STATE $\textbf{k} = \texttt{randperm}(N)$;
\STATE $l = \texttt{round}(N*\eta)$;
\STATE $\tilde{\textbf{Y}}_L^{(s)} =\tilde{\textbf{Y}}(:,\textbf{k}(1:l)) \in {\mathbb{R}^{l\times C}}$;
\STATE $\tilde{\textbf{Y}}_U^{(s)} =0$;
\STATE $\tilde{\textbf{Y}}_{LU}^{(s)} =[\tilde{\textbf{Y}}_L^{(s)}; \tilde{\textbf{Y}}_U^{(s)}]$;
\STATE $\tilde{\textbf{F}}^{*(s)} = (1 - \alpha )(\textbf{I}-\textbf{T})^{-1}{\tilde{\textbf{Y}}_{LU}}^{(s)}$;
\FOR{$i$ = 1 to $N$}
\STATE ${y}_i^{(s)} = \mathop {\arg \max }\limits_j {\textbf{F}_{ij}^{*(s)}}$
\ENDFOR
\ENDFOR
\FOR{$i$ = 1 to $N$}
\STATE ${y}_i^* = MVA(\{{y}_i^{(1)},{y}_i^{(2)}, \cdots, {y}_i^{(s)}\})$
\ENDFOR
\end{algorithmic}
 \label{algorithmic1}
\end{algorithm}

In the same manner as the label propagation algorithm (LPA) \cite{Kothari2002Learning}, in this paper we present to iteratively propagate the labels of the labeled subset $\tilde{\textbf{Y}}_L$ to the remaining unlabeled subset $\mathcal{X}_U$ based on the spectral-spatial affinity graph. Let ${\textbf{F}=}[ {{\textbf{f}}_1}, {{\textbf{f}}_2} \cdots {\rm{,}} {{\textbf{f}}_N}] \in {\mathbb{R}^{N\times C}}$ be the predicted label. At each propagation step, we expect that each pixel absorbs a fraction of label information from its neighbors within the homogeneous region on the spectral-spatial constraint graph, and retains some label information of its initial label. Therefore, the label of ${\textbf{x}}_i$ at time $t+1$ becomes,
\begin{equation}\label{eq:propagation}
\textbf{f}_i^{t + 1} = \alpha \sum\limits_{{\textbf{x}}_i,{\textbf{x}}_j \in {\mathscr{X}_k}} {{\textbf{T}_{ij}}\textbf{f}_j^t}  + (1 - \alpha )\tilde{\textbf{y}}_{i}^{LU},
\end{equation}
where $0 < \alpha < 1$ is a parameter that balancing the contribution between the current label information and the label information received from its neighbors, and $\tilde{\textbf{y}}_i^{LU}$ is the $i$-th column of $\tilde{\textbf{Y}}_{LU} = [\tilde{\textbf{Y}}_L; \tilde{\textbf{Y}}_U]$. It is worth noting that we set the initial labels of these unlabeled samples as $\tilde{\textbf{Y}}_U =\textbf{0}$.

Mathematically, Eq. (\ref{eq:propagation}) can be also rewritten as follows,
\begin{equation}\label{eq:propagationMatrix}
\textbf{F}^{t + 1} = \alpha {{{\textbf{T}}}\textbf{F}^t}  + (1 - \alpha ){\tilde{\textbf{Y}}_{LU}}.
\end{equation}
Following \cite{zhu2002learning}, we learn that Eq. (\ref{eq:propagationMatrix}) can be converged to an optimal solution:
\begin{equation}\label{eq:solution}
\textbf{F}^{*} = \mathop {\lim }\limits_{t \to \infty }\textbf{F}^{t} = (1 - \alpha )(\textbf{I}-\textbf{T})^{-1}{\tilde{\textbf{Y}}_{LU}}.
\end{equation}
$\textbf{F}^{*}$ can be seen as a function that assigns labels for each pixel,
\begin{equation}\label{eq:label}
{y}_i = \mathop {\arg \max }\limits_j {\textbf{F}_{ij}^*}
\end{equation}

Since the initial label and unlabeled samples are generated randomly,
We can repeat the above process of random assignment (of ``clean'' labeled samples and unlabeled samples) and propagation, and obtain multiple labels for each training sample. In particular, we can get different label matrices ${\tilde{\textbf{Y}}_{LU}}^{(s)}$ at the $s$ th round, $s=1,2,...,S$. Here, $S$ is the total number in iterations. We can then calculate the label assignment matrix $\textbf{F}^{*(1)},\textbf{F}^{*(2)}, \cdots,\textbf{F}^{*(S)}$ according to Eq. (\ref{eq:solution}). Thus, we obtain $S$ labels for $\textbf{x}_i$, ${y}^{(1)},{y}^{(2)}, \cdots,{y}^{(S)}$. The final propagated label can be calculated by MVA~\cite{freund1995boosting}.

Because we fully considered the spatial information of hyperspectral images in the process of propagation, we can expect that these propagated label results are better than the original noisy labels in the sense of the proportion that noisy label samples is decreasing (as the number of iterations increase). We illustrate this point in Fig. \ref{fig:NNLS}, which plots the number of noisy label samples according to the iterations of the proposed RLPA under three different noise levels. With the increase of iteration, the number of noisy label samples becomes less and less. The red dashed line shows the initial number of noisy label samples. Obviously, after a certain number of iterations, the number of noisy label samples is significantly reduced. In our experiments, we fix the value of $S$ to 100.

Algorithm \ref{algorithmic1} shows the entire process of our proposed RLPA based label cleansing method. $MVA$ represents the majority vote algorithm that returns the majority of a sequence of elements.


\begin{figure}[t]
\centering
\centerline{\includegraphics[width=7.0cm]{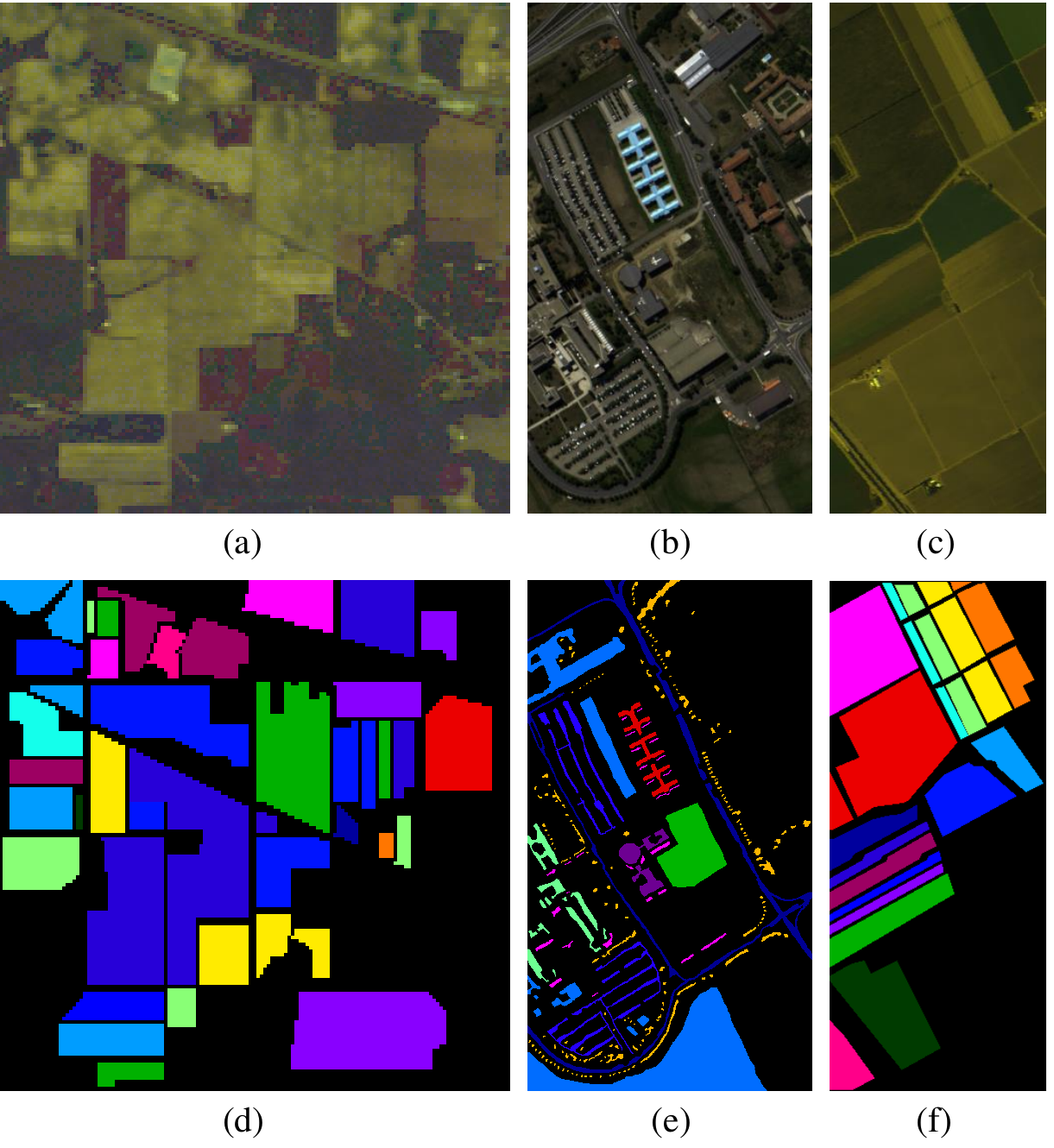}}
\vspace{-0.150cm}
\caption{The RGB composite images and ground reference information of three hyperspectral image databases: (a) Indian Pines, (b) University of Pavia, and (c) Salinas Scene.}
\label{fig:threedatabases}
\end{figure}

\section{Experiments}
\label{sec:Experiments}
In this section, we describe how we set up the experiments. Firstly, we introduce the three hyperspectral image databases used in our experiments. Then, we show the comparison of our results with four other methods. Subsequently, we demonstrate the effectiveness of our proposed SSPTM. Finally, we assess the influence of parameter settings.
We intend to release our codes to the research community to reproduce our experimental results and learn more details of our proposed method from them.

\subsection{Database}
In order to evaluate the proposed RLPA method, we use three publicly available hyperspectral image databases\footnote{\url{http://www.ehu.eus/ccwintco/index.php/Hyperspectral_Remote_Sensing_Scenes}}.
\begin{enumerate}
  \item The first hyperspectral image database is the \emph{Indian Pine}, covering the agricultural fields with regular geometry, was acquired by the AVIRIS sensor in June 1992. The scene is 145$\times$145 pixels with 20 m spatial resolution and 220 bands in the 0.4-2.45 ¦Ìm region. In this paper, 20 low SNR bands are removed and a total of 200 bands are used for classification. \textcolor[rgb]{0.00,0.00,0.00}{This database contains 16 different land-cover types, and approximately 10,249 labeled pixels are from the ground-truth map}.
      Fig.~\ref{fig:threedatabases} (a) shows an infrared color composite image and Fig.~\ref{fig:threedatabases} (d) is the ground reference data.
   \item The second hyperspectral image database is the \emph{University of Pavia}, covering an urban area with some buildings and large meadows, which contains a spatial coverage of 610$\times$340 pixels and is collected by the ROSIS sensor under the HySens project managed by DLR (the German Aerospace Agency). It generates 115 spectral bands, of which 12 noisy and water-bands are removed. It has a spectral coverage from 0.43-0.86 $\mu$m and a spatial resolution of 1.3 m. Approximately 42,776 labeled pixels with nine classes are from the ground truth map, details of which are provided in Table \ref{table:Three_sample}.
       Fig.~\ref{fig:threedatabases} (b) shows an infrared color composite image and Fig.~\ref{fig:threedatabases} (e) is the ground reference data.
  \item The third hyperspectral image database is the \emph{Salinas Scene}, capturing an area over Salinas Valley, CA, USA, was collected by the 224-band AVIRIS sensor over Salinas Valley, California. It generates 512$\times$217 pixels and 204 bands over 0.4-2.5 $\mu$m with spatial resolution of 3.7 m, of which 20 water absorption bands are removed before classification. \textcolor[rgb]{0.00,0.00,0.00}{In this image, there are approximately 54,129 labeled pixels with 16 classes sampled from the ground truth map, details of which are provided in Table \ref{table:Three_sample}}.
      Fig.~\ref{fig:threedatabases} (c) shows an infrared color composite image and Fig.~\ref{fig:threedatabases} (f) is the ground reference data.
\end{enumerate}

\begin{figure*}
\centering
\centerline{\includegraphics[width=16.50cm]{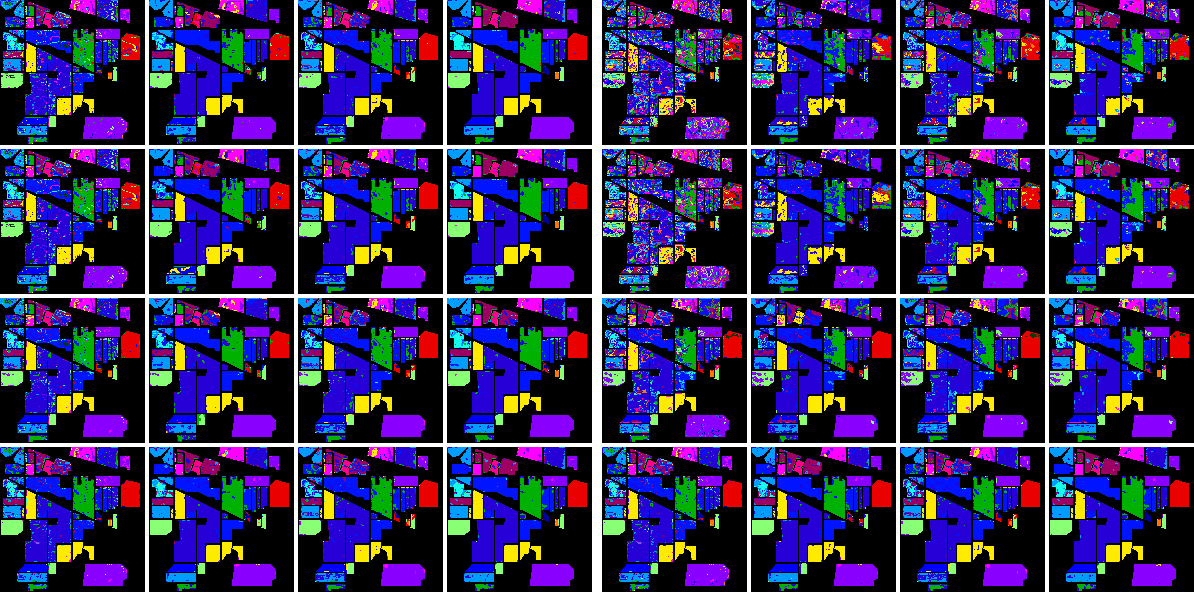}}
\scriptsize {\text{(a) $\rho = 0.1$ {\kern 195pt}  (b) $\rho = 0.5$}}
\vspace{-0.10cm}
\caption{The classification maps of four different methods (each row represents different methods) with four different classifiers (each column represents different classifiers) on the Indian Pines database when (a) $\rho = 0.1$ and (b) $\rho = 0.5$. From the first row to the last row: LNA, Bagging, iForest, and RLPA, from the first column to the last column: NN, SVM, RF, and ELM. Please zoom in on the electronic version to see a more obvious contrast.}
\label{fig:ClassificationMapIndian}
\end{figure*}

For the three databases, the training and testing samples are randomly selected from the available ground truth maps. The class-specific numbers of labeled samples are shown in Table~\ref{table:Three_sample}. For the Indian Pines database, 10\% of the samples are randomly selected for training, and the rest is used testing. As for the other databases, \emph{i.e.}, University of Pavia and Salinas Scene, we randomly choose 50 samples from each class to build the training set, leaving the remaining samples form the testing set.

As discussed previously, we add random noise to the labels of training samples with the level of $\rho$. In other words, each label in the training set will flip to another with the probability of $\rho$. In our experiments, we only show the comparison results of different methods with $\rho \le 0.5$. That is, given a labeled training database, we assume that more than half of the labels are correct, because that the label information is provided by an expert and the labels are not random. Therefore, there are reasons to make such an assumption. Specifically, in our experiments we test typical cases where $\rho = 0.1, 0.2, 0.3, 0.4, 0.5$.

\begin{figure*}
\centering
\centerline{\includegraphics[width=16.50cm]{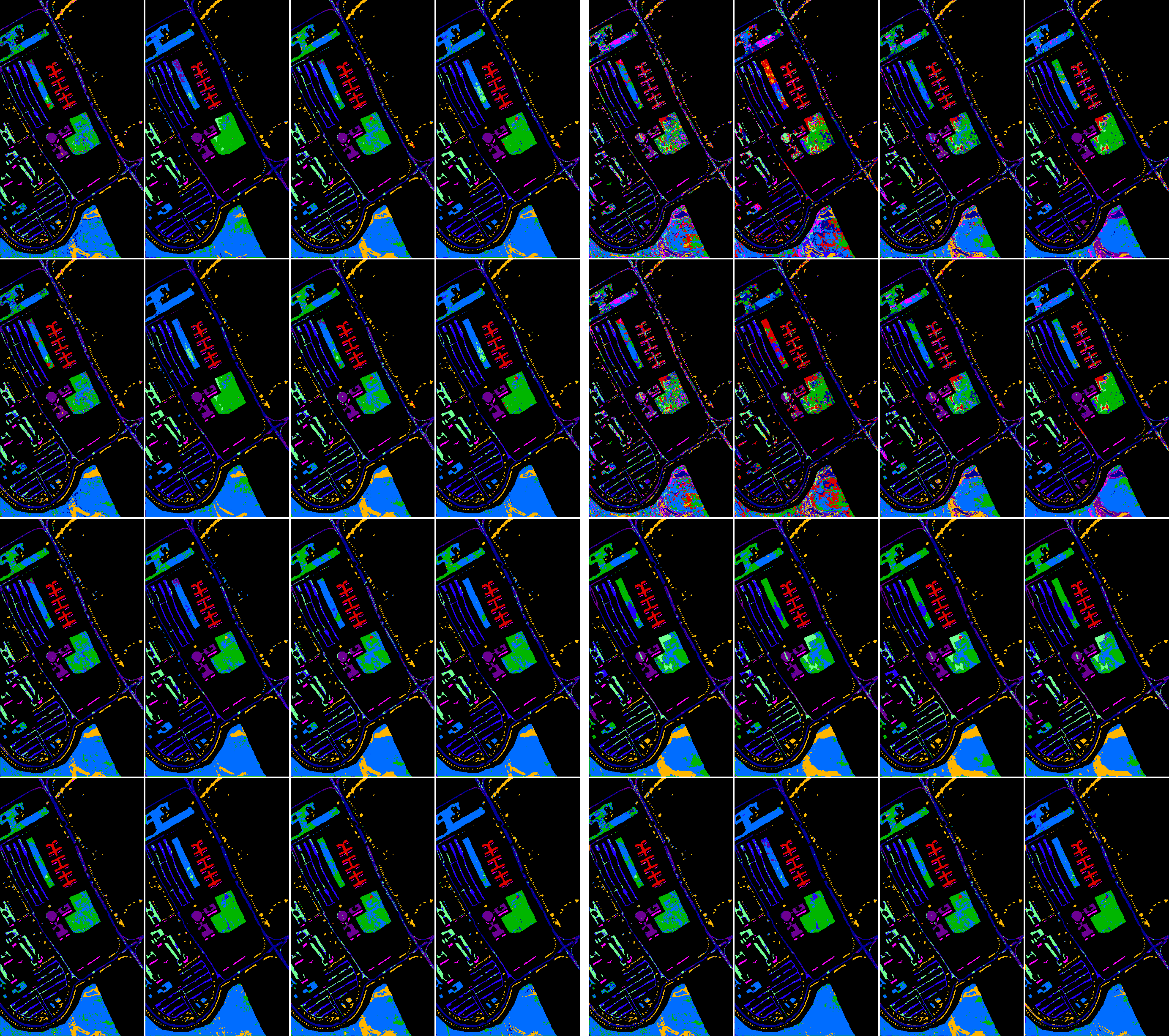}}
\scriptsize {\text{(a) $\rho = 0.1$ {\kern 195pt}  (b) $\rho = 0.5$}}
\vspace{-0.10cm}
\caption{The classification maps of four different method (each row represents different methods) with four different classifiers (each column represents different classifiers) on the University of Pavia database when (a) $\rho = 0.1$ and (b) $\rho = 0.5$. From the first row to the last row: LNA, Bagging, iForest, and RLPA, from the first column to the last column: NN, SVM, RF, and ELM.}
\label{fig:ClassificationMapPaviaU}
\end{figure*}

\subsection{Result Comparison}
\label{sec:comparison}
To demonstrate the effectiveness of the proposed method, we test our proposed framework with four widely used classifiers in the field of hyperspectral image calcification, which are nearest neighborhood (NN) \cite{cover1967nearest}, support vector machine (SVM) \cite{melgani2004classification}, random forest \cite{ham2005investigation, xia2014hyperspectral}, and extreme learning machine (ELM) \cite{li2015local, samat2014rm}. Since there is no specific noisy label classification algorithm for hyperspectral images
, we carefully design and adjust some label noise robust general classification methods to adapt our framework. In particular, the four comparison methods used in our experiments are the following:
\begin{itemize}
  \item Noisy label based algorithm (\emph{NLA}): we directly use the training samples and their corresponding noisy labels to train the classification models using the above-mentioned four classifiers.
  \item Bagging-based classification (\emph{Bagging}) \cite{abellan2012bagging}: the approach of \cite{abellan2012bagging} first produces different training subsets by resampling (70\% of training samples are selected each time), and then fuses the classification results of different training subsets.
  \item isolation Forest (\emph{iForest}) \cite{liu2012isolation}: this is an anomaly detection algorithm, and we apply it to detect the noisy label samples. In particular, in the training phase, it constructs many isolation trees using sub-samples of the given training samples. In the evaluation phase, the isolation trees can be used to calculate the score for each sample to determine the anomaly points. Finally, these samples will be removed when their anomaly scores exceeds the predefined threshold.
  \item \emph{RLPA}: the proposed random label propagation based label noise cleansing method operates by repeating the random assignment and label propagation, and fusing the label information by different iterations.
\end{itemize}

\begin{figure*}[!t]
\centering
\centerline{\includegraphics[width=16.50cm]{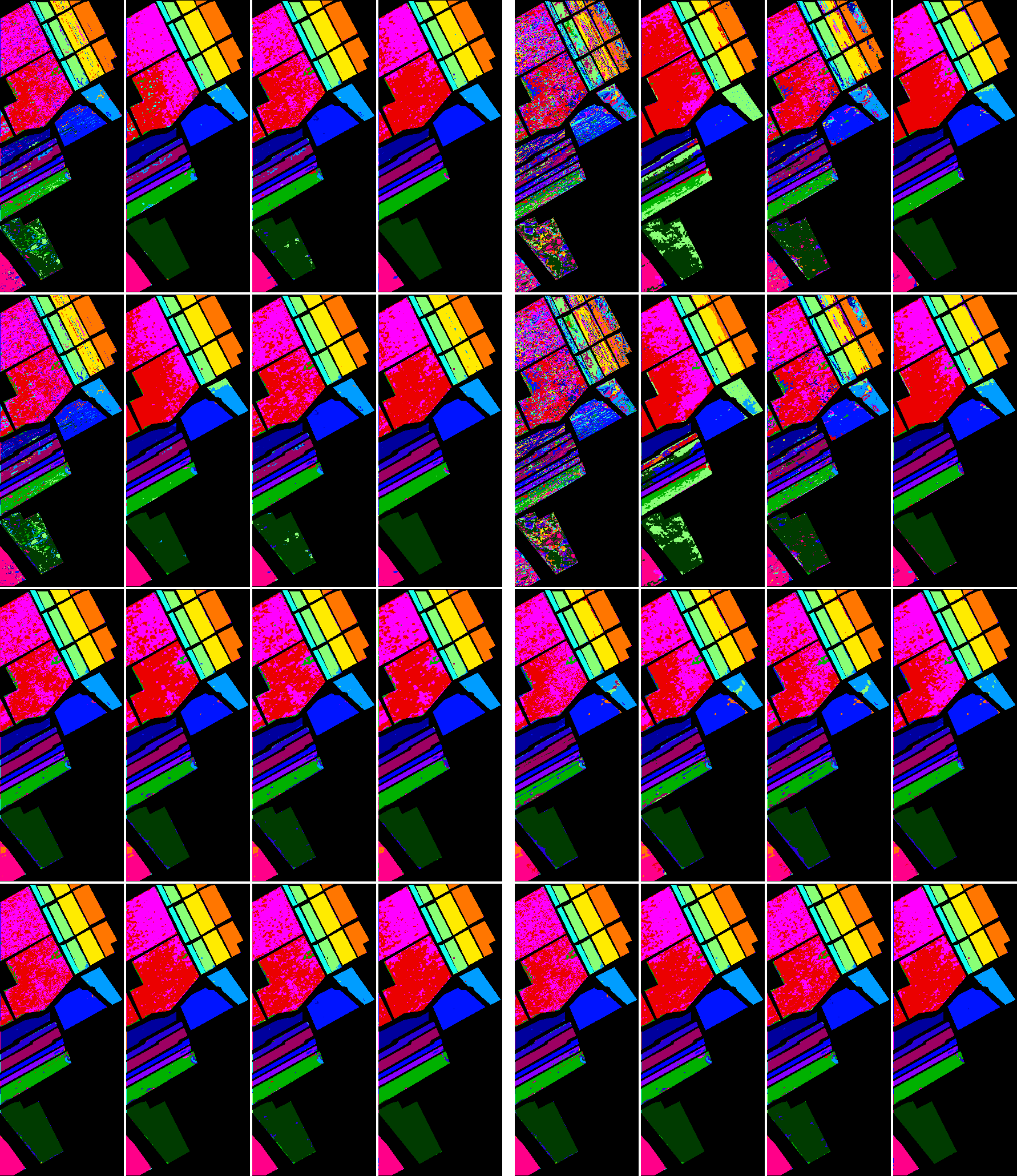}}
\scriptsize {\text{(a) $\rho = 0.1$ {\kern 195pt}  (b) $\rho = 0.5$}}
\vspace{-0.10cm}
\caption{The classification maps of four different methods (each row represents different methods) with four different classifiers (each column represents different classifiers) on the Salinas Scene database when (a) $\rho = 0.1$ and (b) $\rho = 0.5$. From the first row to the last row: LNA, Bagging, iForest, and RLPA, from the first column to the last column: NN, SVM, RF, and ELM.}
\label{fig:ClassificationMapSalinas}
\end{figure*}

\begin{table*}
\centering
\caption{\textcolor[rgb]{0.00,0.00,0.00}{OA, AA, and Kappa performance of four different methods with four different classifiers on the Indian Pines database}.}
\includegraphics[width=13.2cm]{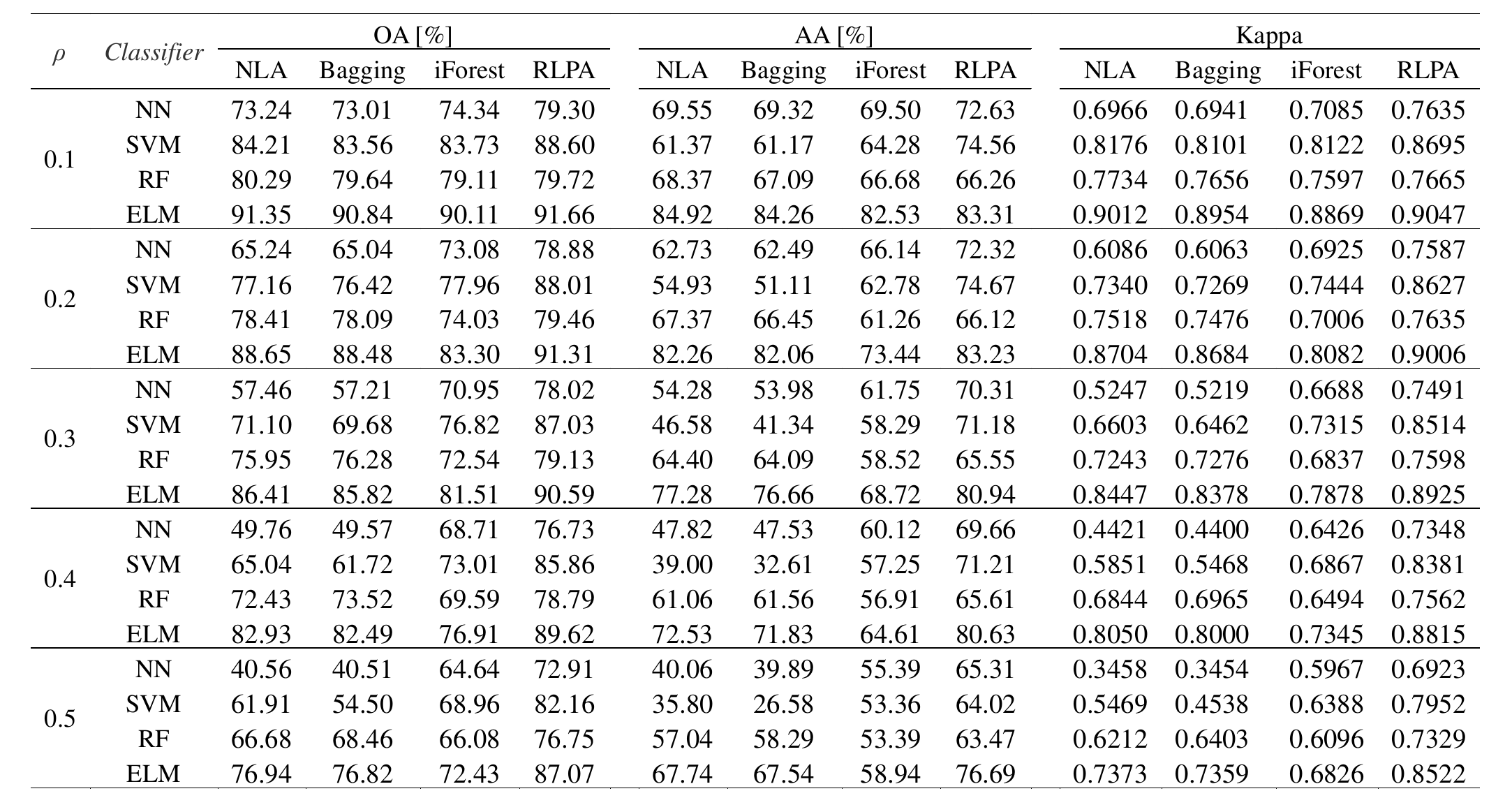}\\
\label{tab:Indian}
\caption{\textcolor[rgb]{0.00,0.00,0.00}{OA, AA, and Kappa performance of four different methods with four different classifiers on the University of Pavia database}.}
\includegraphics[width=13.2cm]{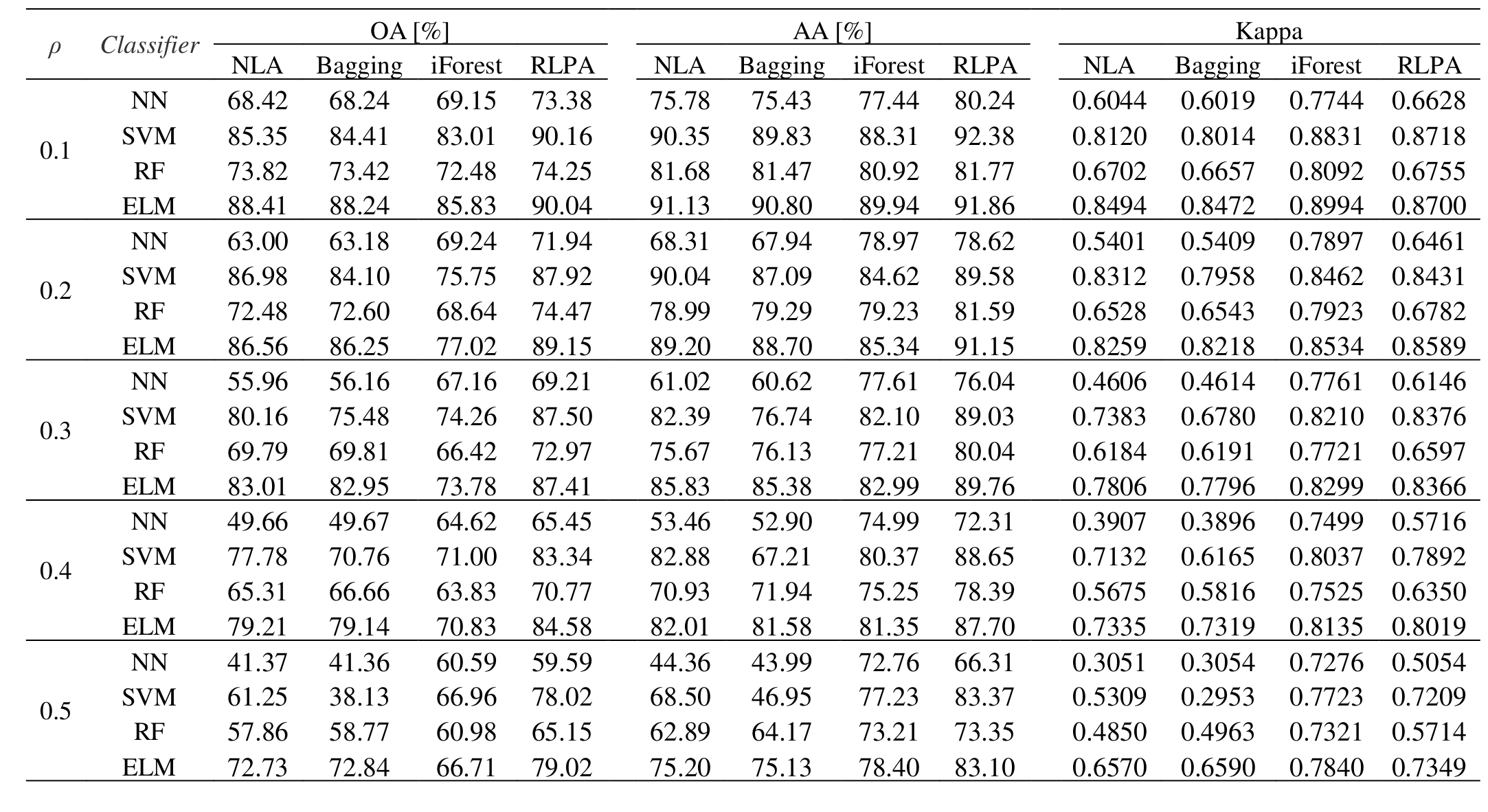}\\
\label{tab:PaviaU}
\caption{\textcolor[rgb]{0.00,0.00,0.00}{OA, AA, and Kappa performance of four different methods with four different classifiers on the Salinas Scene database}.}
\includegraphics[width=13.2cm]{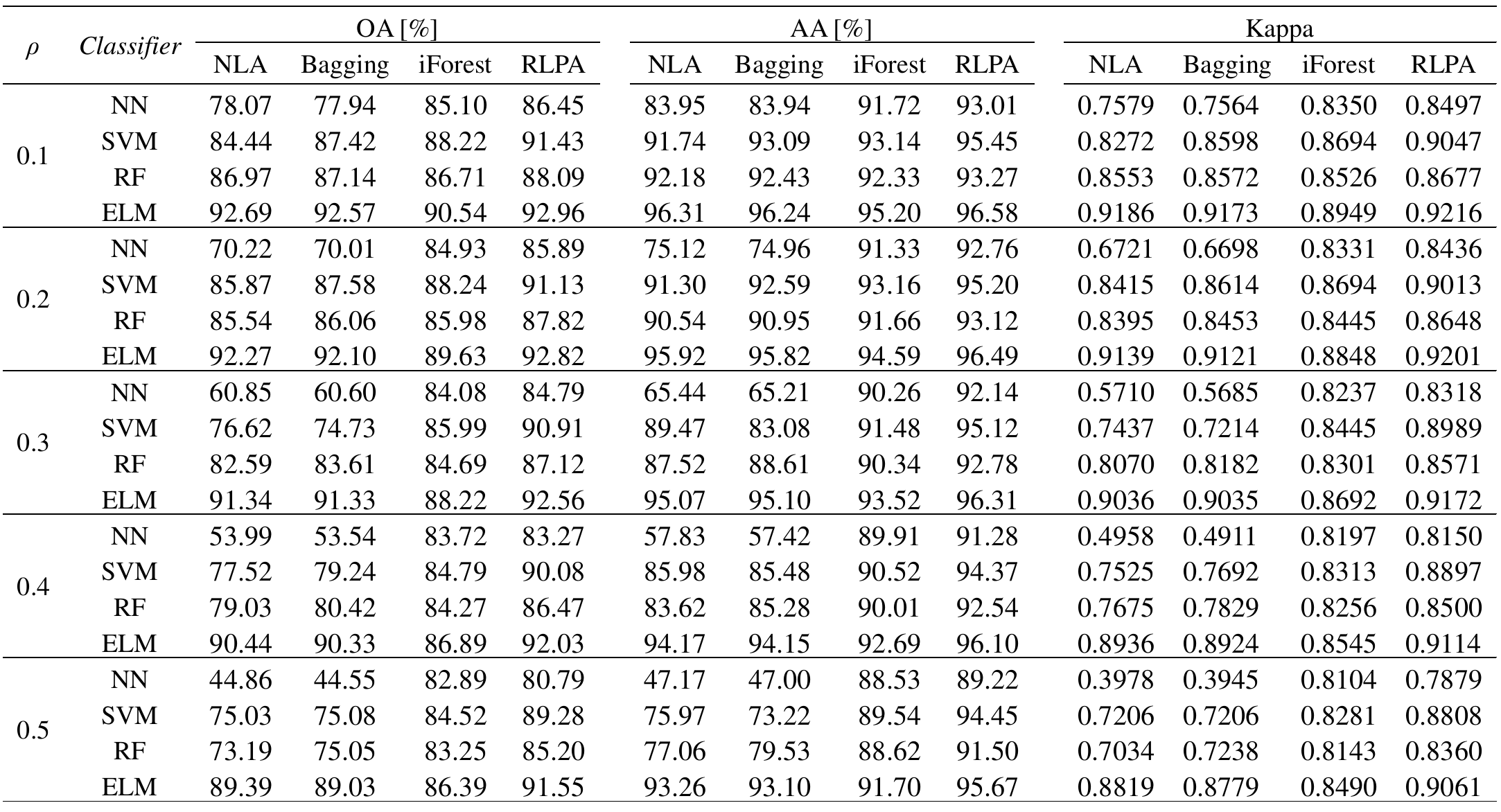}\\
\label{tab:Salinas}

\end{table*}

NLA can be seen as a baseline, Bagging-based method \cite{abellan2012bagging} is a classification ensemble strategy that has been proven to be robust to label noise \cite{krieger2001boosting}. iForest \cite{liu2012isolation} can be regarded as a label cleansing processing as our proposed method in the sense that the goals of these methods are to remove the samples with noisy labels. \textcolor[rgb]{0.00,0.00,0.00}{In our experiments, we carefully tuned the parameters of the four classifiers to achieve the best performance under different comparison methods. Specifically, set all parameters to a larger range, and the reported results of different comparison methods with different classifiers are the best when setting appropriate values for the parameters.}

Generally speaking, the OA, AA, and the Kappa coefficient can be used to measure the performance of different classification results. \textcolor[rgb]{0.00,0.00,0.00}{In Table \ref{tab:Indian}, Table \ref{tab:PaviaU}, and Table \ref{tab:Salinas}, we report the OA, AA, and Kappa scores of four different methods with four different classifiers on the Indian Pines, University of Pavia, and Salinas Scene databases, resepctively. The average OA, AA, and Kappa of LNA, Bagging, iForest, and RLPA for all cases are reported at Table \ref{tab:averages}.  To make the comparison more intuitive, we plot their OA performance in Fig. \ref{fig:Improvement}\footnote{Since these three measurements of OA, AA, and Kappa are consistent with each other, we only plot the results in terms of OA in all our experiments}. In the legend of each subfigure, we also give the average OA of all five noise levels of different methods. From these results, we can draw the following conclusions:}

\begin{table}[t]
\centering
 \caption{\textcolor[rgb]{0.00,0.00,0.00}{The average performance in terms of OA, AA, and Kappa of NLA, Bagging, iForest, and RLPA}.}\label{tab:averages}
 \begin{tabular}{|c||c|c|c|}
\hline
Methods	&	OA [\%]	&	AA [\%]	&	Kappa	\\
\hline
NLA	&	73.93	&	73.26	&	0.6951	\\
Bagging	&	73.20	&	71.94	&	0.6865	\\
iForest	&	77.09	&	78.04	&	0.7293	\\
RLPA	&	83.11	&	82.84	&	0.7994	\\
 \hline
 \end{tabular}
\end{table}

\begin{itemize}
  \item When compared with using the original training samples with label noise (\emph{i.e.}, the NLA method), the Bagging method cannot boost the performance. This indicates that re-sampling the training samples cannot improve the performance of the algorithm in the presence of noisy labels. Moreover, the strategy of re-sampling will result in decreasing the total amount of training samples, so that the classification performance may also be degraded, \emph{e.g.}, the performance of Bagging is even worse than NLA.
  \item The performance of iForest (the cyan lines) is classifier and database dependent. Specifically, it performs well on the NN for all three databases, but it may be even worse than the NLA and Bagging methods. From the average result, iForest can gain more than three percentages when compare to NLA. It should be noted that as an anomaly detection algorithm, iForest has a bottleneck that it can only detect the noise samples but cannot cleanse its label.
  \item The proposed RLPA method (the red lines) can obtain better performance (especially when the noise level is large) than all comparison methods in almost all situations. The improvement also depends on the classifier, \emph{e.g.}, the gain of RLPA over NLA can reach 10\% for the NN and SVM classifiers and will reduce to 3\% for the RF and ELM classifiers. Nevertheless, the gains in term of the average OA, AA, Kappa of our proposed RLPA method over the NLA are still very impressive, \emph{e.g.}, 9.18\%, 9.58\%, and 0.1043.
\end{itemize}

To further demonstrate the classification results of different methods, in Fig. \ref{fig:ClassificationMapIndian}, Fig. \ref{fig:ClassificationMapPaviaU}, and Fig. \ref{fig:ClassificationMapSalinas}, we show the visual results in term of the classification map on two noise levels ($\rho=0.1$ and $\rho=0.5$) for the three databases. For each subfigure, each row represents different methods and each column represents different classifiers. Specifically, from the first row to the last row: LNA, Bagging, iForest, and RLPA, from the first column to the last column: NN, SVM, RF, and ELM. When compared with LNA, Bagging, and iForest, the proposed RLPA with ELM classifier achieves the best performance. However, the classification maps of RLPA may result in a salt-and-pepper effect especially in the smooth regions, whose pixels should be the same class. This is mainly because that the RLPA is essentially a pixel-wise method, and the neighbor pixels may produce inconsistent classification results. To alleviate this problem, the approach of incorporating spatial constraint to fuse the classification result of RLPA can be expected to obtain satisfying results.

\subsection{Effectiveness of SSPTM}
\begin{figure}
\centering
\centerline{\includegraphics[width=8.80cm]{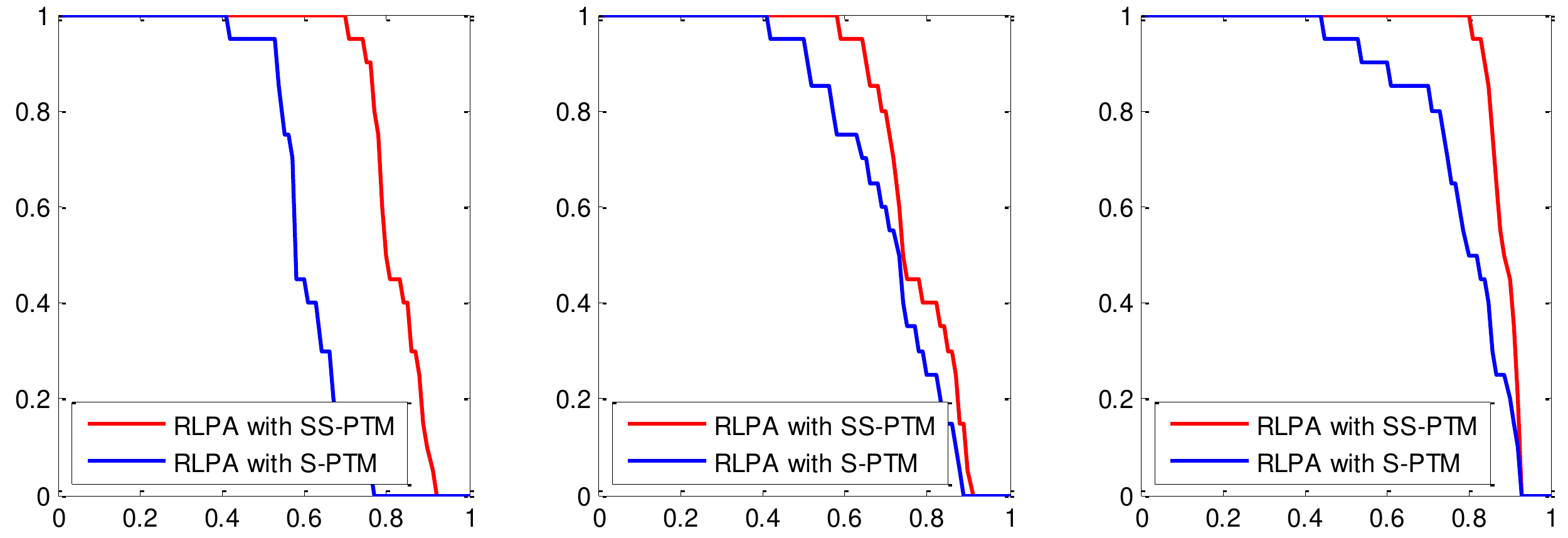}}
\footnotesize {\text{{\kern 0pt} (a) Indian Pines {\kern 20pt} (b) University of Pavia {\kern 23pt} (c) Salinas}}
\vspace{-0.10cm}
\caption{Classification accuracy statistics using RLPA with/without spatial constraint on the three databases. The horizontal axis represents the OA scores, while the vertical axis marks the percentage of larger than the score marked on the horizontal axis.}
\label{fig:SSPTM}
\end{figure}

To verify the effectiveness of the proposed spectral-spatial probability transform matrix generation method, we compare it to the baseline that the similarity between two pixels in only calculated by their spectral difference. To compare the results of spectral-spatial probability transform matrix (SS-PTM) based method and spectral probability transform matrix based method (S-PTM), in Fig. \ref{fig:SSPTM}, we report the statistical curves of OA scores of four comparison methods with four classifiers, \emph{i.e.}, a vector containing 20 elements, whose values are the OA of different situations. It shows a considerable quantitative advantage of SS-PTM compared to S-PTM.

\begin{figure}
\centering
\centerline{\includegraphics[width=8.00cm]{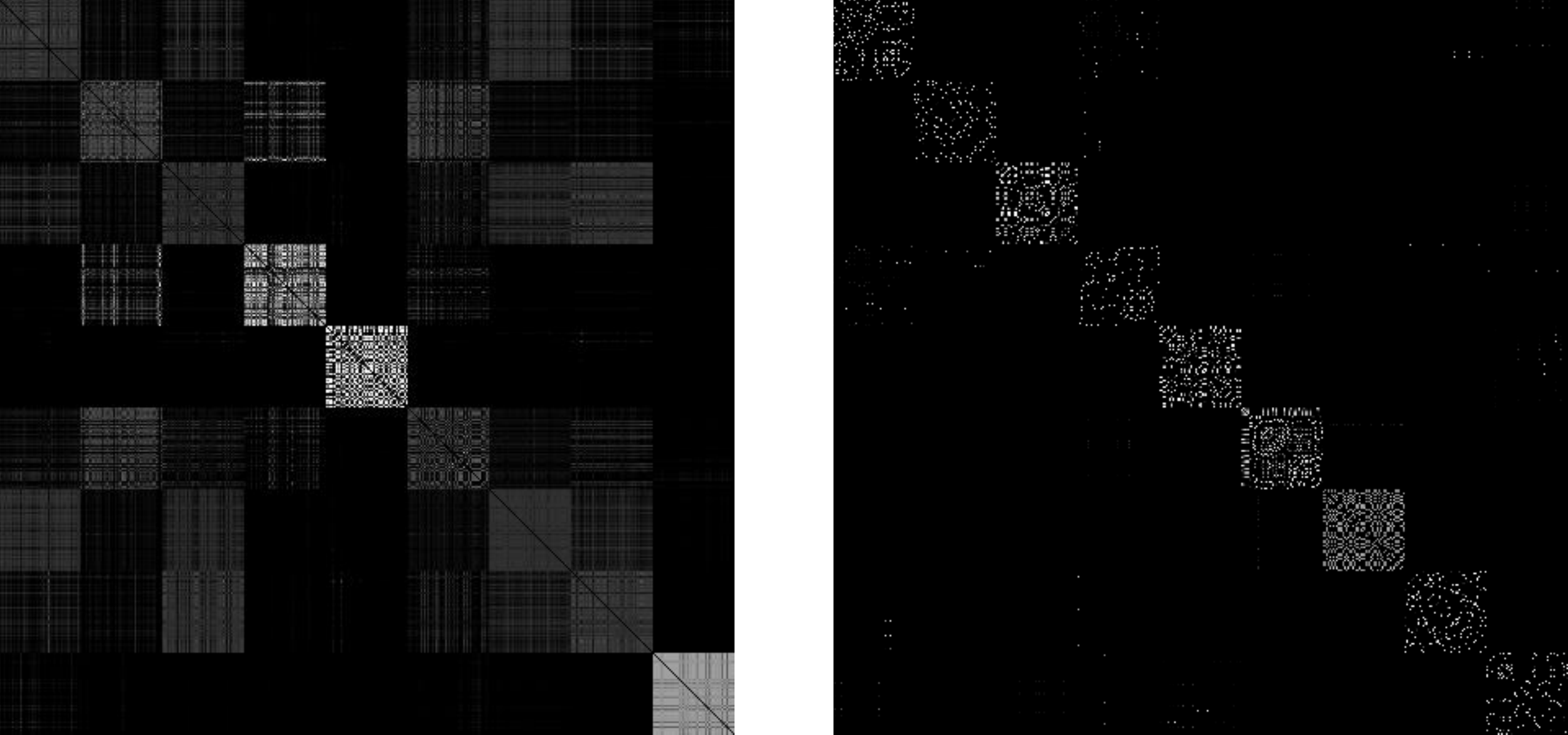}}
\footnotesize {\text{(a) S-PTM {\kern 85pt} (b) SS-PTM}}
\vspace{-0.10cm}
\caption{Visualizations of the probability transfer matrices of (a) S-PTM and (b) SS-PTM. Note that we rescale the intensity values of the matrix for observation.}
\label{fig:SSTMSimMatrix}
\end{figure}

To further analysis the effectiveness of introducing the spatial constraint, in Fig. \ref{fig:SSTMSimMatrix} we show the probability transform matrices with/without a spatial constraint. The two matrices are generated on the University of Pavia database, in which includes 9 classes and 50 training samples per class. From the results, we observe that SS-PTM is a sparse and highly diagonalization matrix, and S-PTM is a dense and non-diagonal matrix. That is to say, SS-PTM does make sense for recovering the hidden structure of data and guarantees the label propagation only within the same class. In contrast to S-PTM, which has many edges between samples with different labels (please refer to the non-diagonal blocks), it may wrongly propagate the label information.

\subsection{Parameter Analysis}
\begin{figure}[h]
\centering
\centerline{\includegraphics[width=7.30cm]{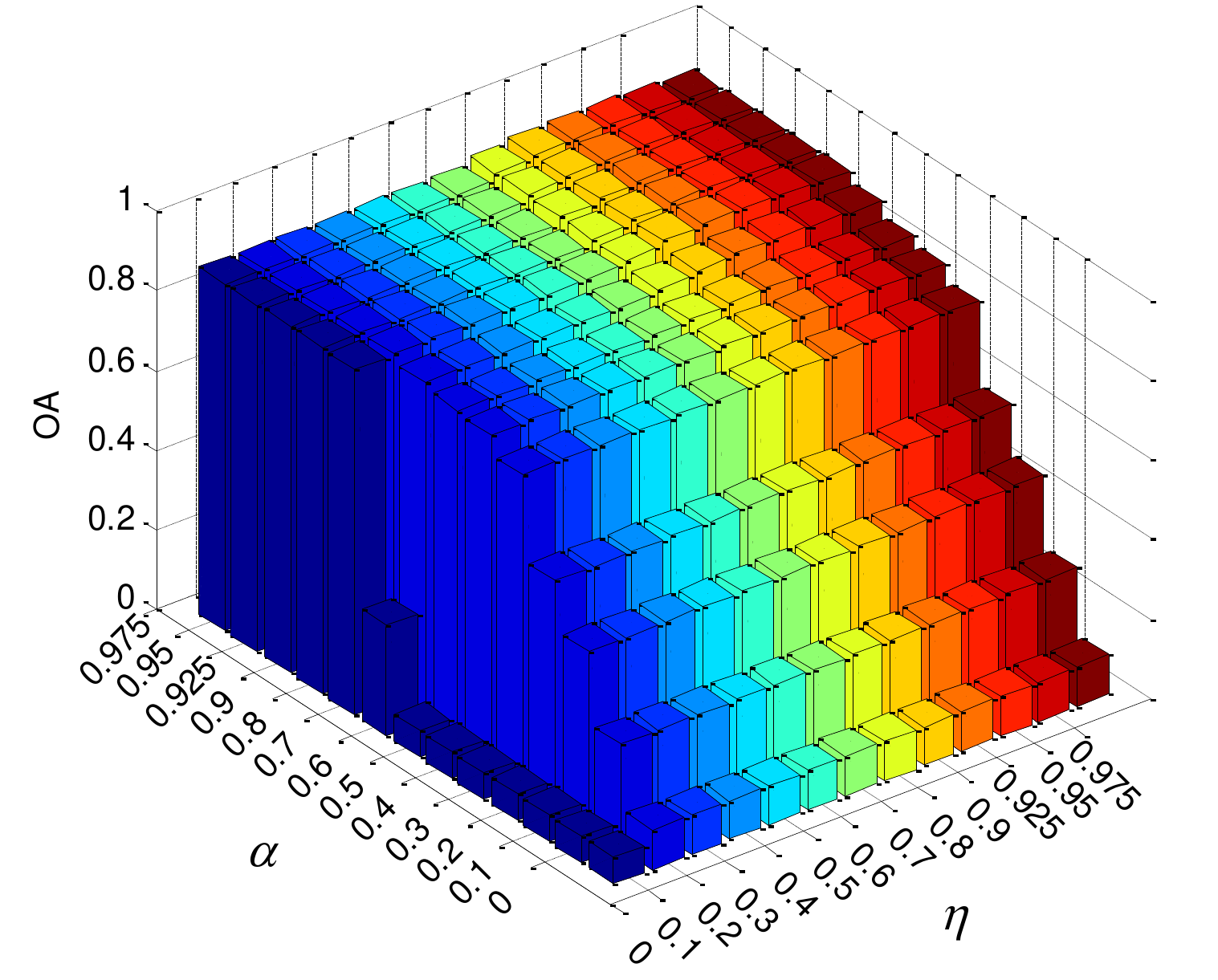}}
\vspace{-0.10cm}
\caption{The classification result in term of average OA on the three databases with ELM classifier under $\rho=0.3$ according to different $\alpha$ and $\eta$, whose values vary from 0.1 to 0.975.}
\label{fig:EtaAlpha}
\end{figure}

From the framework of RLPA, we learn that there are two parameters determining the performance of the proposed method: (i) the parameter $\eta$ denoting the ``clean'' sample proportion in the total training samples, and (ii) the parameter $\alpha$ used to balance the contribution between the current label information and the label information received from its neighbors. In our study, we empirically set their values by grid search. Fig. \ref{fig:EtaAlpha} shows the influence of these two parameters on the classification performance in term of OA. 
It should be noted that we only give the average results of RLPA on the three databases with ELM classifier under $\rho=0.3$. In fact, we can obtain similar conclusions under other situations. From the results, we observe that too small values of $\eta$ or $\alpha$ may be inappropriate. This indicates that ``clean'' labeled samples play an important role in label propagation. If too few ``clean'' labeled samples are selected ($\eta$  is small), the label information will be insufficient for the subsequent effective label propagation process. At the same time, as the value of $\eta$ becomes larger, the performance also starts to deteriorate. This is mainly because that too large value of $\eta$ will make the label propagation meaningless in the sense that very few samples need to absorb label information from its neighbors. In our experiments, we fix $\eta$ to 0.7. Similarly, the value of $\alpha$ cannot be set too large or too small. A too small value of $\alpha$ implies that the final labels completely determined by the selected ``clean'' labeled samples. At the same time, a too large value of $\alpha$ will make it very difficult to absorb label information from the labeled samples. In our experiments, we fix $\alpha$ to 0.9.

\section{Conclusion And Future Work}
\label{sec:Conclusion}
In this paper we study a very important but pervasive problem in practice---hyperspectral image classification in the presence of noisy labels.
The existing classifiers assume, without exception, that the label of a sample is completely clean. However, due to the lack of information, the subjectivity of human judgment or human mistakes, label noise inevitably exists in the generated hyperspectral image data. Such noisy labels will mislead the classifier training and severely decrease the classification performance. Therefore, in this paper we develop a label noise cleansing algorithm based on the random label propagation algorithm (RLPA). RLPA can incorporate the spectral-spatial prior to guide the propagation process of label information. Extensive experiments on three public databases are presented to verify the effectiveness of our proposed approach, and the experimental results demonstrate much improvement over the approach of directly using the noisy samples.

In this paper, we simply use random noise to generate noisy labels. For all classes, they have the same percentage of samples with label noise. However, in real conditions label noise may be sample-dependent, class-dependent, or even adversarial. For example, when the mislabeled pixels come from the edge of the region or are similar to one another, such noise will be more difficult to handle. Therefore, how to deal with real label noise will be our future work.



\ifCLASSOPTIONcaptionsoff
  \newpage
\fi



%
{
\bibliographystyle{IEEEtran}
\bibliography{RLPA2018}
}

\end{document}